\documentclass[a4paper,twoside]{ieeetran}
\usepackage[english]{babel}
\usepackage{tikz}
\usepackage{hyperref}
\usepackage{gensymb}
\usepackage{algorithm}
\usepackage{algpseudocode}
\usepackage{amsmath}
\usepackage{amssymb}
\usepackage{subcaption}
\usepackage{tabularx}
\usepackage{booktabs}
\usepackage{siunitx}
\begin{document}
\title{Autonomous Navigation of Quadrupeds Using Coverage Path Planning with Morphological Skeleton Map}

\author{Alexander James Becoy$^{1,2}$, Kseniia Khomenko$^1$,
        Luka Peternel$^{1,*}$, Raj Thilak Rajan$^2$
\thanks{$^1$ Department of Cognitive Robotics, ME, Delft University of Technology.}
\thanks{$^2$ Department of Microelectronics, EEMCS, Delft University of Technology.}
\thanks{$^*$Corresponding author: Luka Peternel, \href{mailto:l.peternel@tudelft.nl}{l.peternel@tudelft.nl}.}
\thanks{The source code is open source and is available at: \url{https://github.com/asil-lab/go2-autonomous-navigation}}
}

\maketitle
\thispagestyle{plain}
\pagestyle{plain}

\begin{abstract}   

This paper proposes a novel method of coverage path planning for the purpose of scanning an unstructured environment autonomously. The method uses the morphological skeleton of the prior 2D navigation map via SLAM to generate a sequence of points of interest (POIs). This sequence is then ordered to create an optimal path given the robot's current position. To control the high-level operation, a finite state machine is used to switch between two modes: navigating towards a POI using Nav2, and scanning the local surroundings. We validate the method in a leveled indoor obstacle-free non-convex environment on time efficiency and reachability over five trials. The map reader and the path planner can quickly process maps of width and height ranging between [196,225] \SI{}{pixels} and [185,231] \SI{}{pixels} in \SI{2.52}{\milli s} and \SI{1.7}{\milli s}, respectively, where their computation time increases with \SI{22.0}{\nano s/pixel} and \SI{8.17}{\micro s/pixel}, respectively. The robot managed to reach \SI{86.5}{\%} of all waypoints over all five runs. The proposed method suffers from drift occurring in the 2D navigation map.

\end{abstract}

\section{Introduction}

Due to advancements in technology and miniaturization, in the past decade surface (or ground) robots, such as wheeled and legged robots, have been increasingly adopted for diverse operations in harsh and unstructured environments. One of the key challenges in such environments is that the infrastructure to support diverse operations does not readily exist. These environments include, for example, disaster response~\cite{chiou2022robot,lin2022integrated,solmaz2024robust}, mining operations~\cite{parades2021mining,ai2024robotasasensorformingsensingnetwork}, space exploration~\cite{jiang2022space,candalot2024softgrippingspaceexploration,arm2019spacebok,zebro2025}, surveillance in remote locations~\cite{miller2020mine,chagoya2024data}, or hazardous industries like nuclear power plant maintenance~\cite{chen2022nuclear,sharma2024reactors}.

In such complex environments, legged robots are more versatile and robust compared to wheeled robots than other surface robots such as wheeled rovers, and can adaptively navigate uneven, rugged, or soft terrain. Legged robots can cover relatively larger spatial areas by choosing safe footholds within their range of motion and rapidly responding to adjust their kinematic configuration~\cite{yin2023footholds} to achieve their objectives. The number of legs in a legged robot determines its movement efficiency and ability to maintain stability~\cite{nitulescu2016designing}. Compared to bipedal humanoids, quadruped robots demonstrate a greater load capacity and improved stability due to their broader base of support. On the other hand, quadrupeds possess simpler structures and control mechanisms than hexapodal and octopodal robots~\cite{fan2024review, chai2022survey}. For this reason, quadruped robots are ideal for tasks involving safe navigation of complex 3D environments for (sub-)surface exploration.

\begin{figure*}
    \centering
      \begin{subfigure}{0.3\linewidth}
        \centering
        \includegraphics[height=4.5cm]{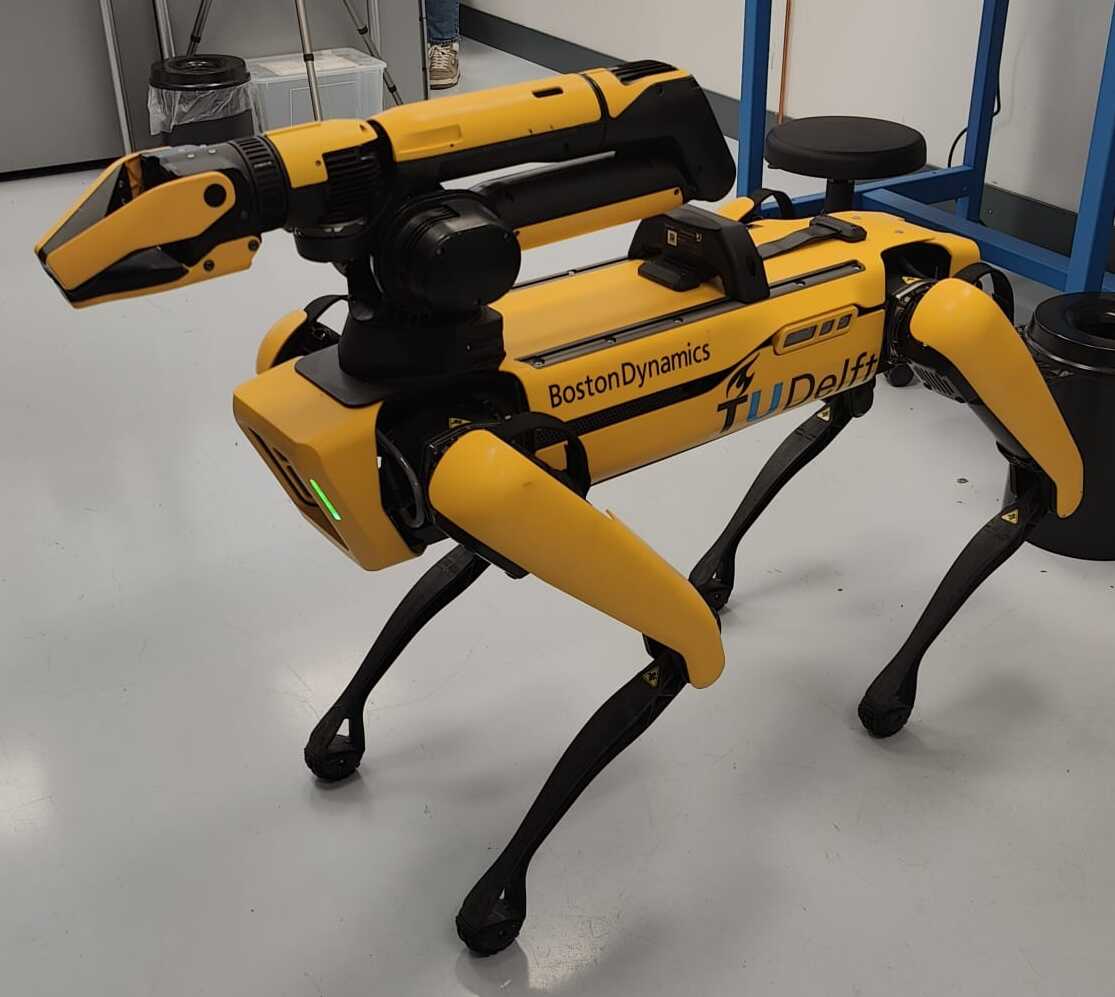}
        \vspace{0.15cm}
        \caption{Boston Dynamics Spot.}
        \label{fig:quadruped_bostondynamics_spot}
    \end{subfigure}
    \hspace{10pt}
    \begin{subfigure}{0.3\linewidth}
        \centering
        \includegraphics[height=4.5cm]{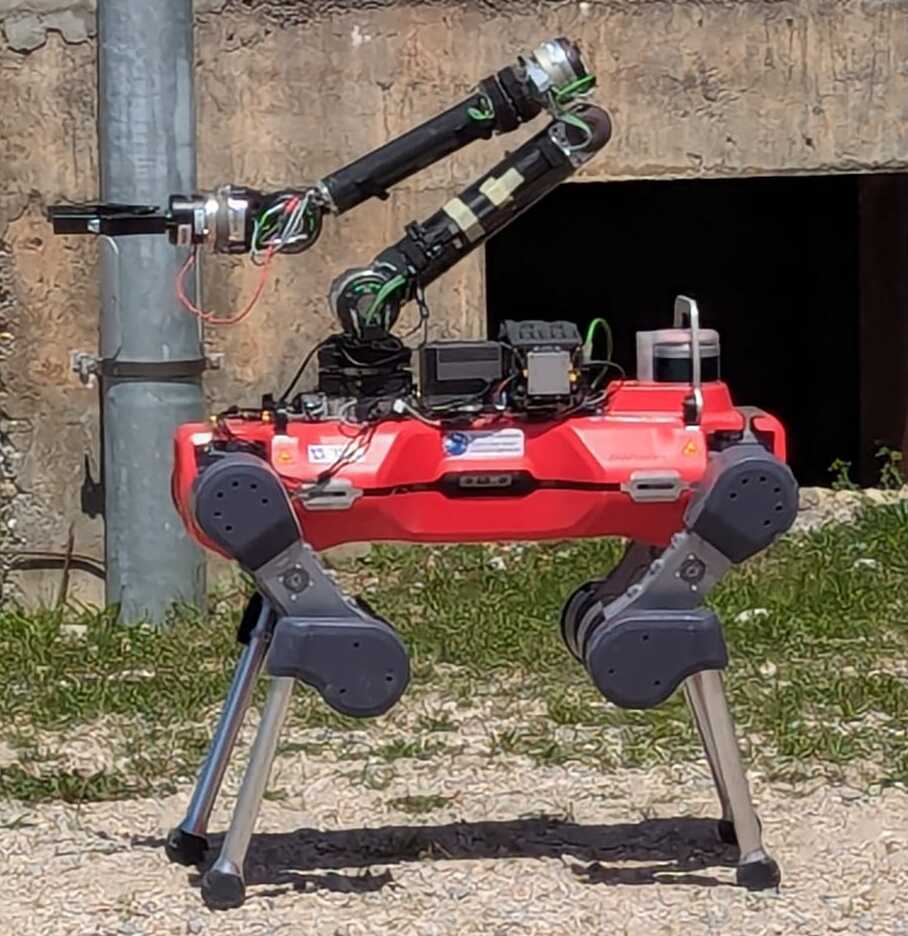}
        \vspace{0.15cm}
        \caption{Anybotics Anymal.}
        \label{fig:quadruped_anybotics_anymal}
    \end{subfigure}
    \hspace{10pt}
    \begin{subfigure}{0.3\linewidth}
        \centering
        \includegraphics[height=4.5cm]{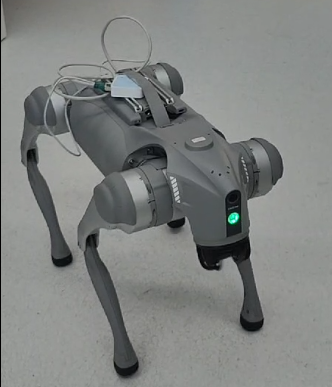}
        \vspace{0.15cm}
        \caption{Unitree Robotic Go2 Edu.}
        \label{fig:quadrupeds_unitree_go2}
    \end{subfigure}
    \caption{Commercially available quadrupedal robots from different companies.}
    \label{fig:quadrupeds}
\end{figure*}

There are several quadrupedal robots that are already commercially available on the market. We compare three notable examples, Boston Dynamics' Spot, ANYbotics' ANYmal, and Unitree Robotic's Go2 Edu as seen in Figure~\ref{fig:quadrupeds}, regarding attributes related to access of development, operation durability, and affordability, as seen in Table~\ref{table:quadruped_comparison}. 
Both Spot and Anymal have garnered significant popularity and have made substantial contributions in research and engineering~\cite{portela2024anymal, zimmermann2021spot} compared to that of Go2 Edu. However, their operational runtime is limited, and their cost is considerably higher. Go2 has three modes: Air, Pro, and Edu with a cost of 1600 USD, 2800 USD, and 12500 USD, respectively. However, only the Edu mode allows for software development, which is necessary for custom implementations, including other necessary features. Furthermore, Go2 Edu has a dedicated Robotics Operating System 2 (ROS 2) integration that allows rapid development and testing. With a factor of two to three in running time, experiments can be done over a long session and the robot can provide sufficient battery capacity to power additional sensors. This paper focuses on the development of a software architecture for path planning and navigation specific to Go2 Edu. 

The goal of this research is to enable quadruped robots to map the terrain using coverage path planning. For quadrupeds to do this, they require sensors, where the most common are cameras \cite{zhang2021objectdetection,zhang2025reidentification,zhang2025benchmark} and LiDAR \cite{bouman2022planning,niu2024skeleton}. To be able to scan with the sensors, the robot needs to move to various POIs, which requires coverage path planning and navigation methods. These will be examined in the related works below.

\begin{table*}[ht]
\caption{Comparison of commercially available quadruped robots.}
\label{table:quadruped_comparison}
\centering
\begin{tabularx}{\textwidth}{llll}
\toprule
\textbf{Feature} & \textbf{Spot} & \textbf{ANYmal D } & \textbf{Go2 Edu} \\
\midrule
\textbf{Manufacturer} & Boston Dynamics & ANYbotics & Unitree Robotics \\
\textbf{Dimensions (L/W/H, mm)} & 1100 x 500 x 191 & 930 x 530 x 890 & 700 x 310 x 400 \\
\textbf{Maximum walking speed (m/s)} & 1.6 & 1.3 & 3.7 \\
\textbf{Average running time (min)} & 90 & 90-120 & 120-240 \\
\textbf{Integrated LiDAR} & No & Yes & Yes \\
\textbf{Integrated optical camera} & Yes & Yes & Yes \\
\textbf{Integrated depth camera} & Yes & Yes & No \\
\textbf{Connectivity} & Wi-Fi, Ethernet & Wi-Fi, 4G & Wi-Fi, Bluetooth, 4G, Ethernet \\
\textbf{Custom Software Development} & Supported (SDK, APIs) & Supported (ROS Integration, APIs) & Supported (SDK, ROS Integration) \\
\textbf{Estimated cost (1000 USD)} & 74.5 & 150 & 12.5 \\
\bottomrule
\end{tabularx}
\end{table*}

\subsection{Related Works}
Several studies have recently explored the development of software infrastructure for the Unitree Robotics Go2 Edu robot. In~\cite{wang2024quadrupedgpt}, a reinforcement learning method was developed to enable the Go2 Edu robot to navigate narrow pipes using visual inputs from a depth camera. The navigation process is relatively straightforward due to the grid-like structure of the pipes, where each pipe is aligned in a straight path with occasional protrusions serving as obstacles.
The work in~\cite{guo2024learning} developed a high-level path planning using a large language model, namely OpenAI's ChatGPT-4o, which allows the interpretation of human verbal commands and translates them into a list of executable instructions. The system integrates a depth camera alongside a segmentation model to perceive the environment effectively. In addition, research in~\cite{cheng2024} developed a motion controller for the Go2 Edu robot to traverse complex and unstructured environments using proprioceptive sensing and collision estimation only.

Autonomous navigation using quadrupedal robots is crucial for exploring complex environments; however, limited work has been done on 2D coverage path planning. Research in~\cite{ly2023planning} achieved coverage path planning for loco-manipulation through an integrated end-to-end pipeline combining perception, optimization, and whole-body motion planning with RGB-D camera inputs. Another work in~\cite{bouman2022planning} presented a 2D coverage path planner for investigating unknown and unstructured environments while accounting for time-bounded and dynamics constraints, and traversability risk. The advantage of these approaches is that they guarantee timely execution when the mission is time-bound, and they find a good optimal tradeoff between maximum area coverage and path traveled. The drawback of that approach is that it is focused on time constraints, therefore, it does not work when the task has a variable execution time and requires time to be defined in advance. In contrast, our approach enables planning for variable times.

Some navigation methods already enable planning for variable times. In the recent work~\cite{niu2024skeleton}, a novel autonomous exploration method was developed using a topological skeleton of the environment's geometry via LiDAR and a finite state machine (FSM) that enables exploratory strategy. This was demonstrated on a quadrupedal robot in an unstructured environment. The advantage of these approaches is very adaptable in an unknown environment due to the use of a state machine that continuously checks for undiscovered areas within the map. The main difference between this work and our approach is that their method for obtaining topological skeletons uses wave propagation and Voronoi diagrams, while our method uses a morphological technique by treating the environment as an image. The impact of this is that the morphological technique simplifies the mapping and is thus computationally more efficient.

Furthermore, all the above methods on coverage path planning have specific map generation algorithms that are an integral part of the whole approach. Therefore, they are less modular, and it is not easy to replace them with developing state-of-the-art mapping algorithms. In contrast, our approach can work with different types of mapping algorithms, thus making it more modular.

\subsection{Contribution}

In this work, we address the gap in the state-of-the-art with the following key contributions.

\begin{enumerate}
    \item We develop a control framework which is based on a finite state machine that can switch between different operation modes that enable autonomous navigation and environmental inspection. 
    \item Using a prior 2D navigation map of the surroundings, the framework rapidly generates an efficient path based on the morphological skeleton of the map, which ensures coverage from a small to a large area.
    \item We validate the developed system with on-field experiments on navigational tasks using the Unitree Robotics Go2 Edu robot in an indoor environment.
    \item We designed an extended interface that enables a Unitree Robotics Go2 Edu to work together with the control framework using ROS 2. 
    \item The whole application is open-source, and it is available at \url{https://github.com/asil-lab/go2-autonomous-navigation}
\end{enumerate}

\section{Preliminaries}
\label{chapter:background}

\subsection{Hardware Specifications}
\label{section:robot}

The Unitree Robotics Go2 Edu features three degrees of freedom (DOFs) per leg, consisting of hip, thigh, and calf hinge joints (from base to foot). It is equipped with an inertial measurement unit (IMU), a HD wide-angle camera, and foot-end force sensors. The robot offers a battery life of 2 to 4 hours and supports fast charging \footnote{Go2 SDK Development Guide - About Go2 \url{https://support.unitree.com/home/en/developer/about\_Go2}}.

For navigation and perception, the Go2 Edu is fitted with the Unitree L1- a 4D LiDAR (3D position + 1D greyscale) based on laser time-of-flight (TOF), mounted on its mouth. This LiDAR provides a 360° × 90° field of view (FOV), a measurement accuracy of ±2.0 cm, and a scanning distance of up to 30 m with 90\% reflectivity. It integrates an IMU with a 3-axis accelerometer and 3-axis gyroscope, has a proximal blind spot of 0.05 m, and features a sampling frequency of 43,200 points per second. Additionally, it operates with a circumferential scanning frequency of 11 Hz and a vertical scanning frequency of 180 Hz \footnote{4D Lidar L1 Application Scenarios 4D Lidar L1 Efficacy — Unitree Robotic \url{https://www.unitree.com/LiDAR}}.
The integrated LiDAR sensor is also utilized for collision detection which facilitates the differentiation between free and occupied spaces in the environment with respect to its height. Additionally, it enables the generation of a 2D map for navigation during run-time, and self-localization ability with respect to environmental features and the current state of the map.

Furthermore, the robot includes an expansion dock, which houses an NVIDIA Jetson Orin with a computing power of 40–100 TOPS. It also comes with a manual two-handed joystick controller for user operation. Additionally, in order to monitor the robot remotely, a TP-Link TL-WR802N Nano WLAN Router is added to set a wireless connection between the robot and the operator's computer.

Lastly, we also integrated external sensors on the robot to measure ambiance such as temperature, humidity, and light intensity. The integration of these sensors is discussed in Appendix~\ref{app:sensing_ambiance}, but the usage of these sensors is outside the scope of this work and is hence not detailed in this paper.

\subsection{Software Specifications}

The Unitree Robotics Go2 Edu robot has a dedicated Software Development Kit (SDK) which allows custom implementations to be programmed. This SDK uses a Data Distribution Service (DDS) as the networking middleware, which enables reliable and real-time data exchange between the program and the robot \footnote{Go2 SDK Development Guide - SDK Concepts \url{https://support.unitree.com/home/en/developer/SDK_Concepts}}.

Using DDS, a Robotics Operating System (ROS) application  \cite{macenski2022ros2} is also implemented to facilitate seamless communication between distributed robotic components, ensuring real-time data exchange, scalability, and interoperability across diverse hardware and software platforms.

In particular, we use ROS 2, given the additional benefits as compared to ROS 1  such as decentralization, simplicity, user-friendliness to name a few. We use the ROS 2 version Foxy on the Linux Ubuntu OS version 20.04 to develop our proposed framework, since those specifications are well established in the Unitree Robotics Go2 Edu robot.

\subsection{SLAM}
\label{section:slam}

\begin{figure}[t]
    \centering
    \includegraphics[width=0.8\linewidth]{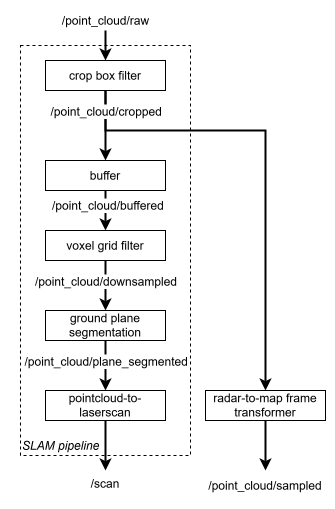}
    \caption{The pipeline diagram of filtering the point cloud for SLAM and 3D scanning per measurement update.}
    \label{fig:pointcloud_pipeline}
\end{figure}

In order for the robot to determine its location within the environment while simultaneously creating a map, we use Simultaneous Localization and Mapping (SLAM). SLAM allows the robot to dynamically create a map based on the history of information, and localize the robot as a function of both the current measurement and the map, simultaneously \cite{stachniss2016slam}. In our case, a 2D map is sufficient because the robot can only navigate on the ground surface in $x$- and $y$-direction (excluding orientation in the $z$-direction). It is worth mentioning that Go2 Edu already comes with its own SLAM implementation. However, at the time of the development, it was not readily available due to lack of documentation. Moreover, we would also like to keep modularity in mind such that this implementation is applicable to other quadrupeds besides the software modules bridging the proposed method and the robot. Therefore we can use Macenski's SLAM Toolbox to create a 2D map using a sequence of 2D laser scans as input \cite{Macenski2021}. This 2D laser scan is the measured distance reflected against an obstacle with respect to the robot's laser scanner at some angle in some given time. Using a radial laser scanner, it can measure the layout of surroundings around the robot simultaneously in one measurement timestep.

Despite Macenski's impressive SLAM implementation, the integrated sensor that measures the geometry of the surroundings is a LiDAR that outputs the measurements as a point cloud in 3D. To make it interpretable for SLAM Toolbox, we process these LiDAR measurements to identify the obstacles which the robot cannot navigate through. To achieve this task, we develop a data processing pipeline using the Point Cloud Library (PCL) \cite{rusu2011pcl} to transform the input 3D point cloud \texttt{/point
\_cloud/raw} in order to obtain the desired output 2D laser scans \texttt{/scan}. An overview of this pipeline is shown in Figure~\ref{fig:pointcloud_pipeline}. Note that this pipeline also outputs another point cloud \texttt{/point\_cloud/sampled}, which will later be used for environment scanning described in Appendix~\ref{app:scanning_procedure}.

\subsection{Navigation}
\label{section:navigation}

We use the Nav2 framework~\cite{macenski2023survey} to ensure the robot is able to plan and navigate towards a desired pose (position and orientation) in the environment, which is also completely compatible with the SLAM Toolbox discussed previously. Using the 2D navigation map provided by SLAM, the Nav2 framework can plan and navigate a path as a function of the destination's pose, the robot's current pose, and the robot's kinematics constraints with respect to its surroundings. A 2D pose is defined as $\boldsymbol{x}:=\begin{bmatrix}x&y&\psi\end{bmatrix}^T$, where $x$ and $y$ define the longitudinal and lateral displacements, and $\psi$ as the orientation of the robot about the $z$-axis (vertical displacement).

The Nav2 framework includes an array of tools such as planners, recoveries and controllers. It is outside the scope of this paper to experiment with different tools. Therefore, we mainly used the default configuration with minor adjustments that corresponds to the robot's kinematics. Due to the fact that the robot's movement can be controlled using 2D velocities $\dot x,\dot y,\dot\theta$ as inputs, we can consider the robot to behave similarly to that of a differential wheeled robot. With this in mind, we can simply use Nav2's default planner \textit{NavFnPlanner}.

Lastly, since the Nav2 framework requires a desired pose as an input, it serves as a local planner and navigation. Therefore, in order to achieve autonomy in the robot, we require a higher-level navigation approach capable of identifying and selecting regions of interest (ROIs) to navigate within the environment.

\section{Methods}
\label{chapter:implementation}

\begin{figure}[t]
    \centering
    \includegraphics[width=\linewidth]{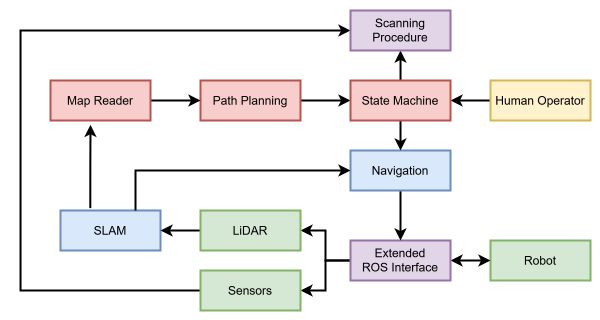}
    \caption{The system overview of the proposed autonomous navigation with scanning capabilities in an unstructured environment. Green: hardware modules, blue: SLAM and Nav2 modules, red: autonomous navigation modules, purple: supplementary modules, and yellow: human agent.}
    \label{fig:implementation_pipeline}
\end{figure}

In this work, we propose a novel framework for the Unitree Robotics Go2 Edu robot for  navigation in unstructured and unpredictable environments. The diagram in Figure~\ref{fig:implementation_pipeline} shows an overview with key modules. The key modules that are responsible for the autonomous navigation are highlighted in red. To create a graph of ROIs, the map reader looks at the SLAM-generated map and transforms it into a topological skeleton based on its geometry (Sec.~\ref{section:map_reader}). The path planning is informed about the graph which creates an efficient path of ROIs, referred to as waypoints, that the robot must take during navigation (Sec.~\ref{section:path_planner}). To control the high-level operation, a state machine (Sec.~\ref{section:state_machine}) is used to switch between actions, e.g., checks when/if each waypoint is complete, if a fallback strategy is needed, if human operator input is detected, etc. The communication between the modules is done via ROS 2 using an extended interface. More information on the extended interface is found in Appendix~\ref{app:extended_ros_interface}.

\subsection{Map Reader}
\label{section:map_reader}

To effectively cover the environment, the robot must follow a path that ensures it traverses every corner of the space. This path should align with the trajectory of the occupied areas, e.g. corridors. This approach is generally effective under the assumption that the environment is confined within a bounded, occupied space, such as an indoor setting. In order to create the path, we process the 2D navigation map as an 8-bit array using the algorithm as seen in Algorithm~\ref{code:map_reader}.  $\mathbf M$, $H$, $W$ denote the map, and the map's height and width, respectively. The map is divided into cells using the map resolution $R$ in m/pixel. Each of these cells is an element in $\mathbf M$, $m_{ij}$, which only contains one of the three types of space: \textit{occupied}, \textit{free}, and $unknown$, which are represented by the integers 0, 255, and 128, respectively.

\begin{algorithm}[t]
\caption{Process the 8-bit 2D navigation map $\mathbf M$ into an unordered set of waypoints $\mathcal{V}$ where each element is a waypoint $\boldsymbol{v}\in\mathbb{R}^2$.}
\label{code:map_reader}
\begin{algorithmic}[1]
\State \textbf{Input:} Navigation map i.e., $\mathbf{M}\in\mathbb{N}^{H\times W}$
\State \textbf{Input:} Hyperparameters  $\sigma,\kappa,\mathbf K, \boldsymbol o\in\mathbb R^2, R$
\State $\mathcal V \leftarrow \emptyset$
\For{$i \in \{1,\ldots,H\}$}
    \For{$j \in \{1,\ldots,W\}$}
        \If{$[\mathbf M]_{ij}<255$}
            \State $[\mathbf M]_{ij} = 0$
        \EndIf
    \EndFor
\EndFor
\State $\mathbf{M}\leftarrow$ GaussianFilter($\mathbf{M}'$, $\sigma$)
\For{$i \in \{1,\ldots,H\}$}
    \For{$j \in \{1,\ldots,W\}$}
        \If{$[\mathbf M]_{ij}>\kappa$}
            \State $[\mathbf M]_{ij} = 1$
        \EndIf
    \EndFor
\EndFor
\State $\mathbf{C}\leftarrow\arg\max_{|\boldsymbol c_i|}$ FindContours($\mathbf{M}$)
\State $\mathbf{M}\leftarrow$ FillAreaByContour($\mathbf{C}$)
\State $\mathbf{M}\leftarrow$ Erode($\mathbf{M}\mid\mathbf{K}$)
\State $\mathbf{M_{\mathbf S}}\leftarrow$ Skeletonize($\mathbf{M}$)
\For{$i \in \{1,\ldots,H\}$}
    \For{$j \in \{1,\ldots,V\}$}
        \If{$[\mathbf M]_{ij}=255$}
            \State $\mathcal V \leftarrow \mathcal V\cup \{R\begin{bmatrix}i&j\end{bmatrix}^T+\boldsymbol{o}\}$
        \EndIf
    \EndFor
\EndFor
\State \textbf{return} $\mathcal{V}$
\end{algorithmic}
\end{algorithm}

First, consider an indoor environment. In such case, most unknown cells inside the map are located outside the boundaries of the occupied area e.g. beyond walls. This is because SLAM initially represents the environment as a grid of unknown cells before any exploration occurs. Furthermore, unknown cells often persist between occupied and free regions as a result of occlusions encountered during raycasting-based measurements, such as those performed using LiDAR, as seen in Figure~\ref{fig:map_reader_result_original}. Consequently, this unknown space cannot be reached by the robot and we can simply treat the unknown regions as part of the boundaries--the occupied space. This can be seen in Figure~\ref{fig:map_reader_result_adjusted}.

To establish a smooth connection between occupied cells affected by noise in the map, a Gaussian filter \cite{dhaeyer1989gaussian} is applied using a standard deviation parameter $\sigma\in\mathbb N$, as seen in Figure~\ref{fig:map_reader_result_fuzzied}. Subsequently, in order to restore the binary representation of occupied and the free cells, the filtered result is binarized using a threshold parameter $\kappa\in[0,255]$. A cell with a value greater than $\kappa$ is classified as a free cell; otherwise, it is designated as an occupied cell.

Due to residual noise and the presence of transparent obstacles, such as windows, this means that there may be free cells that is unreachable to the robot. To mitigate this issue, we assume that the navigable space for the robot corresponds the largest 2D contour. First, the contours within the map are found using the marching cubes algorithm~\cite{cline1987contour}. The map is then reconstructed by filling the largest 2D contour with a value of 255, as seen in Figure~\ref{fig:map_reader_result_contour}. It is worth noting that by only filling the largest contour, it does not take obstacles inside the contour into account.

To ensure that the robot remains a safe distance from the occupied space during navigation, the free space in the map is shrunk using a morphological filter called erosion~\cite{khosravy2017morphology}, as seen in Figure~\ref{fig:map_reader_result_eroded}. Erosion uses a structuring element $\mathbf K$, also referred to as a kernel, that determines the width to be removed. For the sake of simplicity, we take a square matrix of ones as our structuring element: $\mathbf K=\boldsymbol{1}_k\boldsymbol{1}_k^T$, where $k$ determines the length of the vector and $k=1,2,3,\ldots$. The larger the $k$, the larger the safe distance. In addition, the safe distance reduces when $R$ reduces. Therefore, $k\propto R$. This also removes narrow passages that prove impassable for the robot.

The remaining free space is reduced to a thin one-pixel-wide representation containing $O$ pixels, which corresponds to the topological skeleton of the map's geometry. This can be seen in Figure~\ref{fig:map_reader_result_skeleton}. In a practical scenario, the time complexity of 2D skeletonization is mainly proportional to the total number of pixels in an image $P$ as it iterates until the object becomes one-pixel wide. Therefore, it can be considered as $\mathcal O(P)$~\cite{zhang1984thin}. 

The 2D skeleton map $\mathbf{M_{\mathbf S}}$ is then flattened into an unordered set that describe the $x$- and $y$-position of every 2D point by transforming the pixel coordinates into real-world coordinates using the origin $\boldsymbol o$ and resolution $R$ of the 2D navigation map. Only those pixels for which $m_{\mathbf S,ij}=255$ are considered in this transformation. Consider that the skeleton contains $O$ pixels, it follows that there are $O$ waypoints in $\mathcal V$. We refer to every such 2D point defined by the skeleton as \textit{waypoints}, denoted as $\boldsymbol v_i\in\mathbb R^2$ where $0\le i\le O$. The waypoints serve as points of interests for the robot to visit.

\subsection{Path Planning}
\label{section:path_planner}

In order to scan the whole environment, the robot should perform the scanning procedure for each waypoint. Therefore, the objective is to determine a path that includes every waypoint the robot must visit while optimizing for time efficiency. Ultimately, this can be considered as a Travelling Salesman Problem (TSP) which we approach by formulating a fast and efficient path planning algorithm. It takes a connected acyclic graph $\mathcal G$ as input, and outputs a path $\mathcal P$, which is the ordered set of waypoints. In other words, $\mathcal P$ is a sequence of vertices $\boldsymbol v_i$ traversed by the robot. 

In order to construct the graph $\mathcal G$, we treat the unordered set of waypoints $\mathcal V$, obtained from the map reader, as the set of vertices. Each vertex in $\mathcal V$ uniquely corresponds to the coordinate vector of a waypoint, and can, for the sake of simplicity, be denoted as $\boldsymbol v_i\in\mathbb R^2,\;\forall i=1\ldots,O$. Each vertex is then connected to other vertices if they are neighbors around the map's resolution $R$. This should resemble to the skeleton representation of the map, where the connections define the edges $\mathcal E$ of the graph $\mathcal G$.

\begin{algorithm}[t]
    \begin{algorithmic}[1]
        \State \textbf{Input:} Define graph $\mathcal G(\mathcal V,\mathcal E) $
        \State \textbf{Hyperparameters:} $R$, $D$, $\boldsymbol x_0$
        \State $\mathcal V_{\text{leaf,total}}\leftarrow\emptyset$
        \For{$\boldsymbol v_i\in\mathcal V$}
            \If{degree$(\boldsymbol v_i)=1$}
                \State $\mathcal V_{\text{leaf,total}}\leftarrow\mathcal V_{\text{leaf,total}}\cup \{\boldsymbol v_i\}$
            \EndIf
        \EndFor
        \State $\mathcal V_{\text{leaf,visited}}\leftarrow\emptyset$
        \State $\mathcal P\leftarrow\emptyset$
        \State $\boldsymbol v_{\text{leaf,start}}\leftarrow \text{FindNearestLeaf}(\boldsymbol x_{0},\mathcal V_{\text{leaf,total}})$
        \While{$|\mathcal V_{\text{leaf,visited}}| < |\mathcal V_{\text{leaf,total}}|$}
            \State $\mathcal V_{\text{leaf,visited}}\leftarrow\mathcal V_{\text{leaf,visited}}\cup\{\boldsymbol v_{\text{leaf,start}}\}$
            \State $\boldsymbol{v}_{\text{leaf,target}}\leftarrow \text{FindNearestLeaf}(\boldsymbol v_{\text{leaf,start}},\mathcal V_{\text{leaf,total}}\setminus\mathcal V_{\text{leaf,visited}})$
            \State $\mathcal P'\leftarrow \text{FindShortestPath}(\boldsymbol v_{\text{leaf,start}},\boldsymbol v_{\text{leaf,target}}\mid\mathcal G)$
            \For{$\boldsymbol{v}'_i \in \mathcal P'$}
                \If{$\boldsymbol{v}'_i\not\in \mathcal P$}
                    \State $\mathcal P\leftarrow \mathcal P\cup \{\boldsymbol{v}'_i\}$
                \EndIf
            \EndFor
            \State $\boldsymbol v_{\text{leaf,start}}\leftarrow\boldsymbol v_{\text{leaf,target}}$
        \EndWhile
        \State $\mathcal P^*\leftarrow\emptyset$
        \For{$k\leftarrow0$ to $\left \lfloor{\frac{|\mathcal P|-1}{D/R}}\right \rfloor$}
            \State $\mathcal P^*\leftarrow\mathcal P^*\cup [\mathcal P]_{\frac{D}{R}k}$
        \EndFor
        \State \textbf{return} $\mathcal P^*$
    \end{algorithmic}
    \caption{A fast and efficient approach to planning a path $\mathcal P^*$, with $L$ number of vertices, using the unordered set of waypoints obtained from map reader $\mathcal V$, map resolution $R$, waypoint resolution $D$, and the robot's starting 2D position $\boldsymbol x_0$.}
    \label{code:path_planner}
\end{algorithm}

Algorithm~\ref{code:path_planner} outlines the procedure for obtaining an efficient path $\mathcal P^*$ for coverage exploration of the whole environment. First, there are dead-ends found in the skeleton representation of the map. For the robot to cover the whole environment, we can utilize these dead-ends. Each dead-end is identified as a leaf vertex $\boldsymbol v_{\text{leaf}}\in\mathcal V_{\text{leaf}}$ and is typically defined with degree 1. Assuming the graph $\mathcal G$ is connected and acyclic, complete coverage of the map can be achieved by ensuring that the robot visits every leaf vertex in $\mathcal V_{\text{leaf}}$ in sequence. Since all other vertices in the graph lie on the paths connecting the leaves. traversing to each leaf inherently requires passing through the intermediate vertices. As a consequence, visiting all leaf vertices implies that the entire graph has been traversed. 

After identifying all leaf vertices, the leaf vertex closest to the robot's current 2D position $\boldsymbol{x}_0$ is selected as the initial source leaf vertex $\boldsymbol{v}_{\text{leaf,start}}$ using Algorithm~\ref{code:find_nearest_leaf}. For every source leaf vertex $\boldsymbol{v}_{\text{leaf,start}}$, we find the next nearest unvisited leaf vertex as the target $\boldsymbol{v}_{\text{leaf,target}}$ using FindNearestLeaf, as defined in Algorithm~\ref{code:find_nearest_leaf}. We find the shortest path $\mathcal P'$ between the source $\boldsymbol v_{\text{leaf,source}}$ and the target $\boldsymbol v_{\text{leaf,target}}$ given the graph $\mathcal G$. To achieve this, we use the method FindShortestPath which essentially utilizes Dijkstra's algorithm. This algorithm has a time complexity of $\mathcal{O}(|\mathcal E|+|\mathcal V|\log|\mathcal V|)$, where $|\mathcal V|$ and $|\mathcal E|$ denote the total number of vertices and the total number of edges in the graph, respectively~\cite{barbehenn1998dijkstra}.

Once $\mathcal P'$ is found, each vertex $\boldsymbol{v}'_i\in\mathcal P'$ is appended into $\mathcal P$. Unless some vertex $\boldsymbol{v}'_i$ reappears as a path of a different path $\mathcal P'$, revisiting and scanning at this position is unnecessary and should be avoided to conserve time and computational resources. Once iterated over all $\boldsymbol{v}'_i$ in $\mathcal P'$, $\boldsymbol{v}_{\text{leaf,start}}$ is inserted into a set of visited leaf vertices $\mathcal V_{\text{leaf,visited}}$ and $\boldsymbol{v}_{\text{leaf,target}}$ becomes the next source $\boldsymbol{v}_{\text{leaf,start}}$. Not only is the shortest path useful for appending $\boldsymbol{v}_i$ into $\mathcal P$ but also for ensuring a feasible and unobstructed trajectory through occupied spaces.

\begin{algorithm}[tp]
    \begin{algorithmic}[1]
        \State \textbf{Function} FindNearestLeaf 
        \State \textbf{Input:} $\boldsymbol x_0\in\mathbb R^2$, $\mathcal V'_{\text{leaf}}$
        \State $\boldsymbol v_{\text{leaf,nearest}}\leftarrow\boldsymbol 0$
        \State $d\leftarrow\infty$
        \For{$\boldsymbol v_{\text{leaf,i}}\in\mathcal V'_{\text{leaf}}$}
            \If{$|\boldsymbol v_{\text{leaf,i}}-\boldsymbol x_0|<d$}
                \State $\boldsymbol v_{\text{leaf,nearest}}\leftarrow \boldsymbol v_{\text{leaf,i}}$
                \State $d\leftarrow|\boldsymbol v_{\text{leaf,i}}-\boldsymbol x_0|$
            \EndIf
        \EndFor
        \State \textbf{return} $\boldsymbol v_{\text{leaf,nearest}}$
        \State \textbf{End Function}
    \end{algorithmic}
\caption{A method to find the nearest leaf vertex $\boldsymbol v_{\text{leaf,target}}$ from a set of selected leaf nodes $\mathcal V'_{\text{leaf}}$ to a given source $\boldsymbol x_0$ given the Euclidean distances.}
    \label{code:find_nearest_leaf}
\end{algorithm}

This is iterated until all leaf vertices have been added and $\mathcal P$ is complete. Nevertheless, because the original distance between every two waypoints is approximately equal to the map's resolution, this leads to significant time spent on scanning with minimal displacement. Considering that the robot has a large scanning range capability of 30 m, which allows it to scan distances greater than the map resolution, we can increase the distance between every two waypoints to some arbitrary distance $D$ by reducing $P$ by a factor of $D/R$ where $D \ge R$. This is done by selecting every $D/R$-th element in $\mathcal P$.

\subsection{State Machine}
\label{section:state_machine}

\begin{figure}[t!]
    \centering
    \begin{subfigure}{\linewidth}
        \centering
        \includegraphics[width=\linewidth]{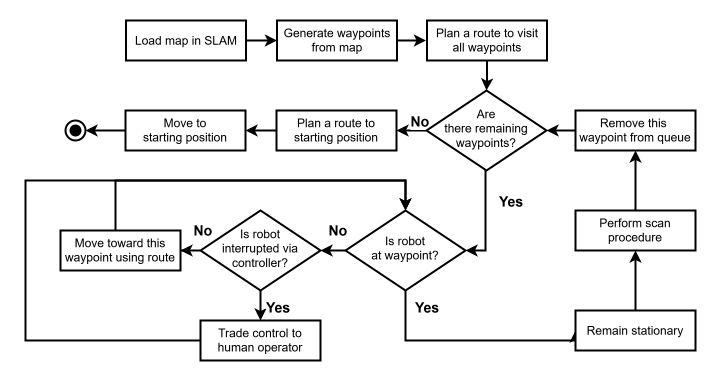}
        \vspace{0.1cm}
        \caption{The activity diagram describing the autonomous coverage path planning given a prior 2D navigation map.\\}
        \label{fig:activity_diagram}
    \end{subfigure}
    \hfill
    \begin{subfigure}{\linewidth}
        \centering
        \includegraphics[width=0.8\linewidth]{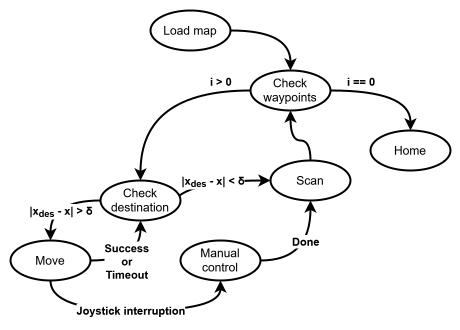}
        \caption{A FSM diagram based on the activity diagram, where the decision arguments $i$ is the number of remaining waypoints to visit, $\mathbf x$ is the robot's current 2D pose, $\mathbf x_{\text{des}}$ is the destination's 2D pose, and $\boldsymbol\delta$  is the predefined tolerance.}
        \label{fig:state_machine}
    \end{subfigure}
    \caption{Comparison of (a) the activity diagram for autonomous coverage path planning and (b) its corresponding FSM diagram.}
    \label{fig:activity_vs_fsm}
\end{figure}

To achieve autonomy that enables the robot to visit the waypoints and perform a scanning procedure in succession, an activity diagram is formulated as seen in Figure~\ref{fig:activity_diagram}. We can group two or more activities as a singular state if the transitions are consecutive without interruptions, e.g., a decision node. This state will perform all activities in their respective order. 
On the other hand, each decision node becomes a state that checks if its control variable has reached a specified threshold. 
Each state has a conditionless trigger that automatically transitions the current state to the next specified state at the end of its action.

With a finite state machine (FSM), we can achieve the desired autonomous navigation based on the triggers by handling the sequence of actions. The implemented state machine diagram can be seen in Figure~\ref{fig:state_machine}. Each state is explained as follows:

\begin{itemize}
    \item In State \texttt{Load Map}, an existing 2D navigation map is loaded onto SLAM and allows the robot to be localized with respect to the map's metadata (meter-per-pixel resolution, width and height both in pixels, and origin as $(x,y)$ in meters), and the robot's concurrent surroundings through 2D laser scans. After this map is loaded, it is then analyzed to generate a list of waypoints, each in $(x, y)$ coordinates, for the robot to visit using the map reader. An optimal route is then planned to visit all of these waypoints, which reorders the original list using the path planner. Finally, it automatically transitions to the next state, \texttt{Check Waypoints}.
    \item State \texttt{Check Waypoints} allows the system to iterate over the list of waypoints. If there are waypoints remaining, it removes the waypoint in the first entry of the list and stores it as the current destination, and transitions to State \texttt{Check Destination}. If there is no waypoint left, it transitions to State \texttt{Home}.
    \item In State \texttt{Check Destination}, the state machine determines if the robot is already at the current destination with some acceptable offset. This is done by determining if the Euclidean distance between the robot's 2D pose $\boldsymbol x$ and a destination's pose $\boldsymbol x_{\text{des}}$, is less than or equal to a set tolerance $\boldsymbol{\delta}$, where $\boldsymbol x,\boldsymbol x_{\text{des}},\boldsymbol{\delta}\in\mathbb R^3$ since the poses are defined in the $x$- and $y$-axes, and the yaw orientation $\psi$ with respect to \texttt{/map} frame. If the condition is true, it transitions to State \texttt{Scan}. Otherwise, it transitions to State \texttt{Move}. It is worth noting that the path planner does not account for the orientation $\psi$. However, this orientation is likely a necessary requirement of the scanning procedure, ensuring that the robot is aligned with the desired orientation, as described in Supplementary Material C.
    \item In State \texttt{Move}, it actuates the robot to navigate towards the desired destination. Navigation is considered successful when the Euclidean distance between the robot’s current pose $\boldsymbol x$ and the target destination $\boldsymbol x_{\text{des}}$ is less than or equal to a predefined threshold $\delta$. Alternatively, if the navigation process exceeds a specified timeout duration $T_{\text{timeout}}$, it is also terminated. In either case, the system transitions back to the \texttt{Check Destination} state. However, if the system detects an interruption by a human operator via the joystick controller mid-operation, the system promptly cancels the ongoing action and immediately transitions to State \texttt{Manual Control}.
    \item In State \texttt{Scan}, the robot performs the procedure to scan the local environment, which is detailed in Supplementary Material C. Once it is completed, it transitions to State \texttt{Check Waypoints}.
    \item State \texttt{Manual Control} allows the human operator to take over the robot's navigation and move it towards the destination. This should happen in a case when the robot is traversing difficult terrains, or when the navigation framework is stuck at finding the right solution. This transitions to State \texttt{Scan} once the operator presses a button on the controller.
    \item In State \texttt{Home}, the robot travels back to its starting position. Once the robot arrives at the starting position, it lies down on the ground and waits for the new commands.
\end{itemize}

To sum up, the proposed framework for navigation consists of three key modules: the map reader to extract POIs using the 2D navigation map as waypoints, the path planner to order the waypoints as an efficient route given the robot's current position, and the state machine to enable the robot to navigate towards every waypoints and scan, consecutively.

\section{Experiments and Results}
\label{chapter:results}

To demonstrate the proposed framework and evaluate how well it performs, we conducted the experiment in a leveled in-door obstacle-free non-convex environment, which emulated an inaccessible environment. For evaluation, we used the following two metrics:

\begin{enumerate}
    \item Time efficiency: the time required to process the map to create a path, and to reach each waypoint consecutively.
    \item Reachability: the number of waypoints that the robot is able to reach over the total number of waypoints planned in \%.
\end{enumerate}

The Go2 Edu robot was initially manually controlled via a wireless controller to generate a 2D navigation map using the SLAM module. The experiment was then conducted and the whole process was repeated over five trials. The SLAM module's map resolution was set to $0.10$ meters. The point cloud buffer time was set to 0.25 seconds. The laser scan's minimal and maximal range was set to 0.50 meters and 30.0 meters, respectively. The smoothing standard deviation was set to 3. The crispification threshold was set to 128. The erosion kernel $K$ was set to 10 x 10 pixels. The waypoint-to-waypoint distance $D$ in path planning was set at 1.00 m. The robot's maximum x-, y- and, yaw velocities were set to 1.00 m/s, 0.50 m/s, and 0.80 rad/s, respectively. The 2D navigation position and orientation tolerance were set to 0.05 m and 0.08 rad, respectively, with a navigation timeout of 10.0 s to replan. The scanning procedure module was disabled to demonstrate the navigational tasks.

\subsection{SLAM, Map Reader and Path Planning}

\begin{figure}[t]
    \centering
    \includegraphics[width=0.65\linewidth]{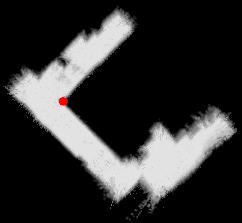}
    \caption{derived by normalizing over the five 2D navigation maps obtained from five trials, where the inner corner of the left wing of the room is taken as the origin which is denoted as a red point. All five maps are translated and rotated around the red point accordingly.}
    \label{fig:slam_result}
\end{figure}

\begin{figure}[t]
    \centering
    \begin{subfigure}[t]{0.45\linewidth}
        \centering
        \includegraphics[width=\linewidth]{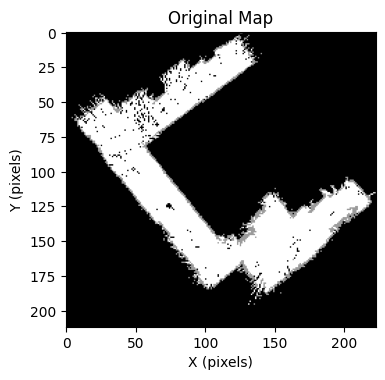}
        \caption{The original map.}
        \label{fig:map_reader_result_original}
    \end{subfigure}
    \begin{subfigure}[t]{0.45\linewidth}
        \centering
        \includegraphics[width=\linewidth]{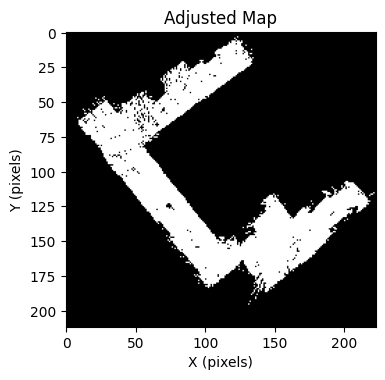}
        \caption{The adjusted map.}
        \label{fig:map_reader_result_adjusted}
    \end{subfigure}
    \begin{subfigure}[t]{0.45\linewidth}
        \centering
        \includegraphics[width=\linewidth]{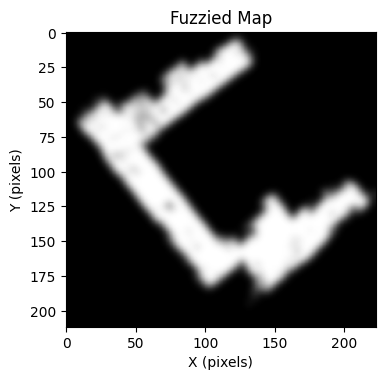}
        \caption{The fuzzied map.}
        \label{fig:map_reader_result_fuzzied}
    \end{subfigure}
    \begin{subfigure}[t]{0.45\linewidth}
        \centering
        \includegraphics[width=\linewidth]{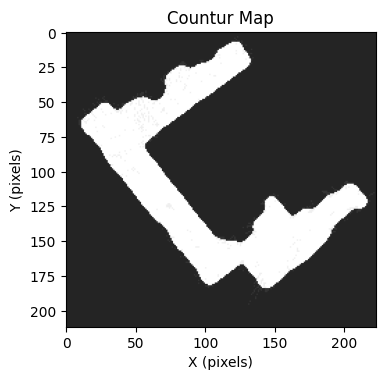}
        \caption{The contour map.}
        \label{fig:map_reader_result_contour}
    \end{subfigure}
    \begin{subfigure}[t]{0.45\linewidth}
        \centering
        \includegraphics[width=\linewidth]{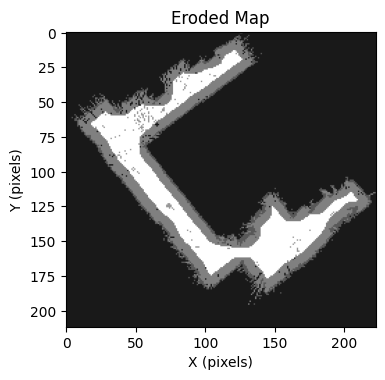}
        \caption{The eroded map.}
        \label{fig:map_reader_result_eroded}
    \end{subfigure}
    \begin{subfigure}[t]{0.45\linewidth}
        \centering
        \includegraphics[width=\linewidth]{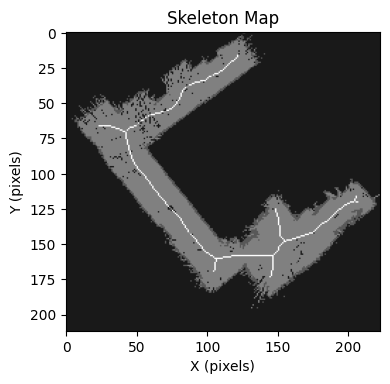}
        \caption{The skeleton map.}
        \label{fig:map_reader_result_skeleton}
    \end{subfigure}
    \caption{A demonstration of the map reader's pipeline using the 2D navigation map of Trial 5 as input.}
    \label{fig:map_reader_result}
\end{figure}

Before presenting the main results, we first validate the accuracy of the 2D navigation maps generated by the SLAM module. The average map is derived by normalizing over the five 2D navigation maps obtained from five trials, as illustrated in Figure~\ref{fig:slam_result}. The width and height of these maps range between $[196,225]$ and $[185,231]$ pixels, respectively. We take the inner corner of the left-wing (upper left triangle of the map) of the room as the origin to align the maps in position and orientation accordingly. Analysis of the averaged map reveals that the right-wing exhibits a wider range of shades of gray about the occupied space, suggesting that this part of the environment is slightly tilted relative to its actual orientation. Nonetheless, this does not severely affect the map reader, path planning, and navigation as the robot mainly localizes itself in its immediate surroundings and the topological skeleton of the map can still be found. For instance, as seen in Figure~\ref{fig:map_reader_result}, the map reader can still create waypoints mainly using the map's geometry where it also enables the robot to visit the corners by producing branches from the main path to these corners.

\begin{figure}[t]
    \centering
    \begin{subfigure}[b]{\linewidth}
        \centering
        \includegraphics[width=0.95\linewidth]{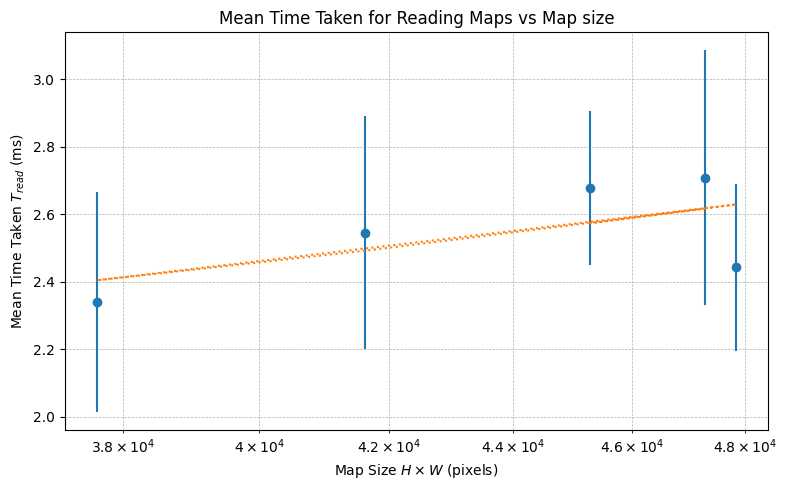}
        \caption{Mean time taken for map reader over five trials, iterated over $N=100$ plotted in blue dots, with standard deviation in 1$\sigma$ as vertical blue line. Trend line is $22.0$ ns/pixel.}
        \label{fig:map_reader_time_taken}
    \end{subfigure}
    \hfill
    \begin{subfigure}[b]{\linewidth}
        \centering
        \includegraphics[width=0.95\linewidth]{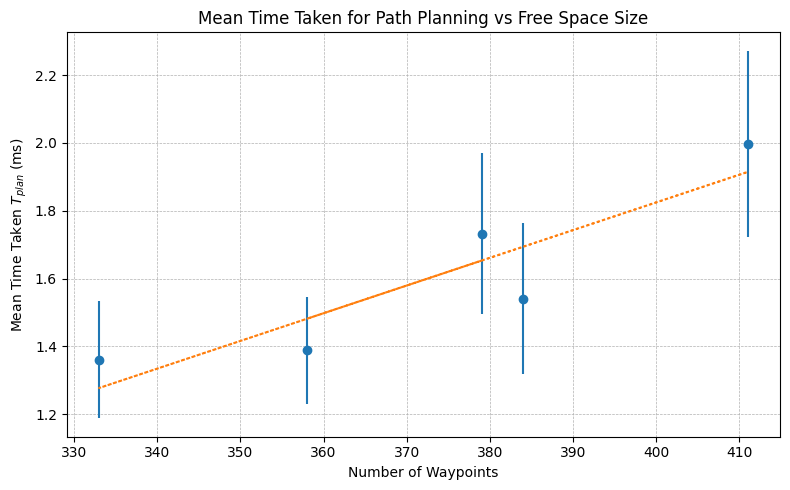}
        \caption{Mean time taken for path planner over five trials, iterated over $N = 500$ plotted in blue dots, with standard deviation in 1$\sigma$ as vertical blue line. Trend line is \SI{8.17}{\micro s/pixel}.}
        \label{fig:path_planner_time_taken}
    \end{subfigure}
    \caption{Comparison of mean time taken for (a) map reader and (b) path planner over five trials.}
    \label{fig:time_taken_comparison}
\end{figure}

To evaluate the time efficiency of the map reader, we record the duration required to process each map across five trials, considering a total of $N=100$ instances. We vary the weights in terms of the dimension per map. To facilitate a clearer representation in the plot, we use the product of the dimensions, $H\times W$ in pixels. As can be seen in Figure~\ref{fig:map_reader_time_taken}, we can see that the average time taken for the map reader ranges from 2.34 ms to 2.70 ms over the five trials with an average standard deviation of about 0.30 ms. We can also see that the mean time increases by 22.0 ns for every additional pixel in the map. For instance, with a map with size $H\times W=10^6$ pixels, it has an expected mean time $T_{\text{read}}\approx23.6$ ms. 

To remain within a maximum time taken of 1.0 s, the map should not be larger than the size of $H\times W=45.4\times10^6$ pixels. Assuming a square map with a map resolution of 0.10 m, this maximum map has a length of 673.4 m. 
Hence, our proposed map reader is fast for relatively small to large map sizes (ranging up to $45.4\times10^6$ pixels) within a lead time of $T_{\text{read}}=1.0$ s, especially when the map reader is only run once to produce the waypoints given the map's geometry.

To determine the path planner's time efficiency, we repeat the same process with the evaluation of the map reader over $N=500$ iterations per trial. Instead, we vary the weights in terms of the number of waypoints per map. In Figure~\ref{fig:path_planner_time_taken}, we can observe that the mean time taken ranges between 1.40 ms and 2.00 ms. The variances can be explained by the additional time required to find the next nearest leaf vertex from a source vertex as it searches in the list sequentially. According to the trend line, the lead time increases with \SI{8.17}{\micro s} for every additional waypoint. Therefore, according to the trend line, it means that for a map with $1000$ waypoints, the planning time becomes $T_{\text{plan}}\approx6.18$ ms, and for a map of $10^6$ waypoints, $T_{\text{plan}}\approx8.17$ s. 

To stay within a lead time of 1.0 s, the number of waypoints should not be larger than $1.219\times 10^5$. The bottleneck appears if the map resolution $R$ becomes small, as it increases the number of pixels per meter in a map. This can substantially increase the number of waypoints by a factor of $R/R'$, where $R'$ is the new map resolution and $R' < R$. Nonetheless, the path planner can be considered fast for a small to large number of waypoints, given a reasonable map resolution $R$. It is also worth noting that it also runs only once to produce the necessary path.

\begin{figure}[t]
    \centering
    \begin{subfigure}[t]{\linewidth}
        \centering
        \includegraphics[width=0.85\linewidth]{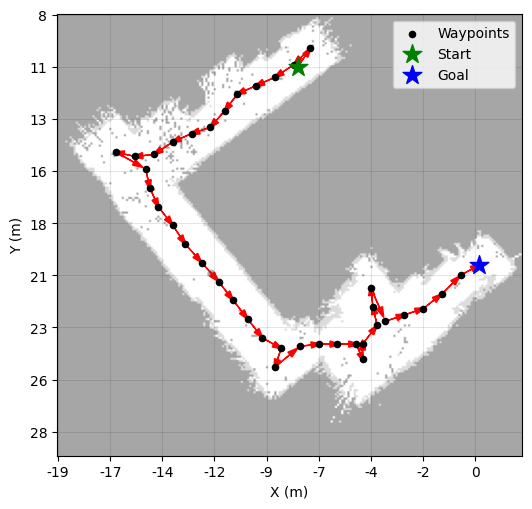}
        \caption{An efficient path to explore the whole environment from the robot's starting position (green) to the farthest waypoint (blue) while visiting all other waypoints (black). Each arrow (red) denotes the direction from source to target.}
        \label{fig:path_planner_result}
    \end{subfigure}
    \hfill
    \begin{subfigure}[t]{\linewidth}
        \centering
        \includegraphics[width=0.85\linewidth]{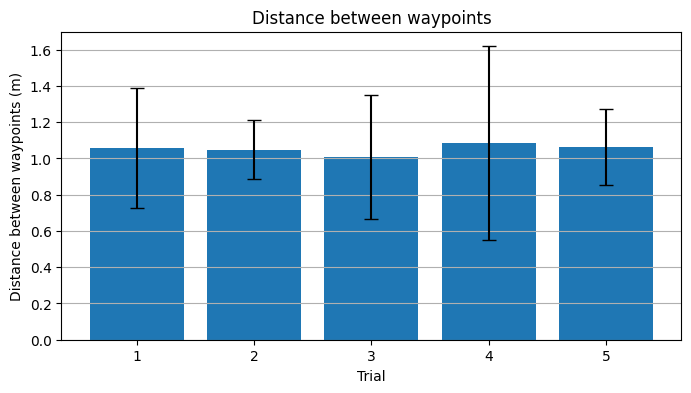}
        \caption{Mean distance between every two connected waypoints generated by the path planning per trial, as defined by the waypoint resolution $D$.}
        \label{fig:waypoint_resolution}
    \end{subfigure}
    \caption{Demonstrations of path planner's performance.}
    \label{fig:path_planning_comparison}
\end{figure}

To verify the path planner's performance, we first evaluate the logic whether it works as intended. Using one trial as an example, we can see that the path planner has created the optimal path with respect to time based on the waypoints given the robot's current 2D position, as seen in Figure~\ref{fig:path_planner_result}. Furthermore, according to Figure~\ref{fig:waypoint_resolution}, we can see that the mean waypoint-waypoint distance is about $1.00$ m given the previously set waypoint resolution $D$. The variance can be explained due to the fact that robot skips already visited waypoints, since it is unnecessary for the robot to scan about the same position twice. This results in the robot having to travel distances of $1.00$ m or more for some intervals. In contrast to traveling longer distance, the robot can also visit the next waypoint in a shorter distance. This is due to the fact that the path splicing happens at the end of the algorithm, whereby resulting in a probability that two waypoints are less than the set waypoint resolution. Nevertheless, this does not heavily affect navigation, and the average waypoint distance is almost identical to the set waypoint resolution $D$.

\subsection{Navigational Performance}

\begin{table*}[t]
\caption{Reachability and time taken for each trial including key observations.}
\label{tab:navigation_performance}
\centering
\begin{tabularx}{\textwidth}{
    >{\centering\arraybackslash}p{0.05\textwidth}
    >{\centering\arraybackslash}p{0.12\textwidth}
    >{\centering\arraybackslash}p{0.1\textwidth}
    >{\centering\arraybackslash}p{0.175\textwidth}
    p{0.45\textwidth}}
\toprule
     \textbf{Trial} & \textbf{Reachability (\%)} & \textbf{Total Time (s)} & \textbf{Median Time per Waypoint (s)} & \textbf{Observations} \\
    \midrule
    1 & 100.0 & 443.2 & 5.40 & Human assistance required at Waypoint 17. \\
    2 & 100.0 & 397.0 & 5.30 & Human assistance required at Waypoint 36. \\
    3 & 82.61 & 287.2 & 5.40 & Significant map drift occurred after 3 minutes, such that the remaining 8 waypoints were unreachable (displaced into the occupied space). \\
    4 & 100.0 & 322.3 & 5.40 & Robot stalled at Waypoints 6, 10, 18, 23, 28, 35, 37-39 due to replanning. \\
    5 & 48.72 & 160.1 & 5.40 & Significant map drift occurred after 2 minutes, such that the remaining 19 waypoints were unreachable as they displaced into the occupied space. \\
\bottomrule
\end{tabularx}
\end{table*}

\begin{figure*}[htb]
    \centering
    \includegraphics[width=0.85\textwidth]{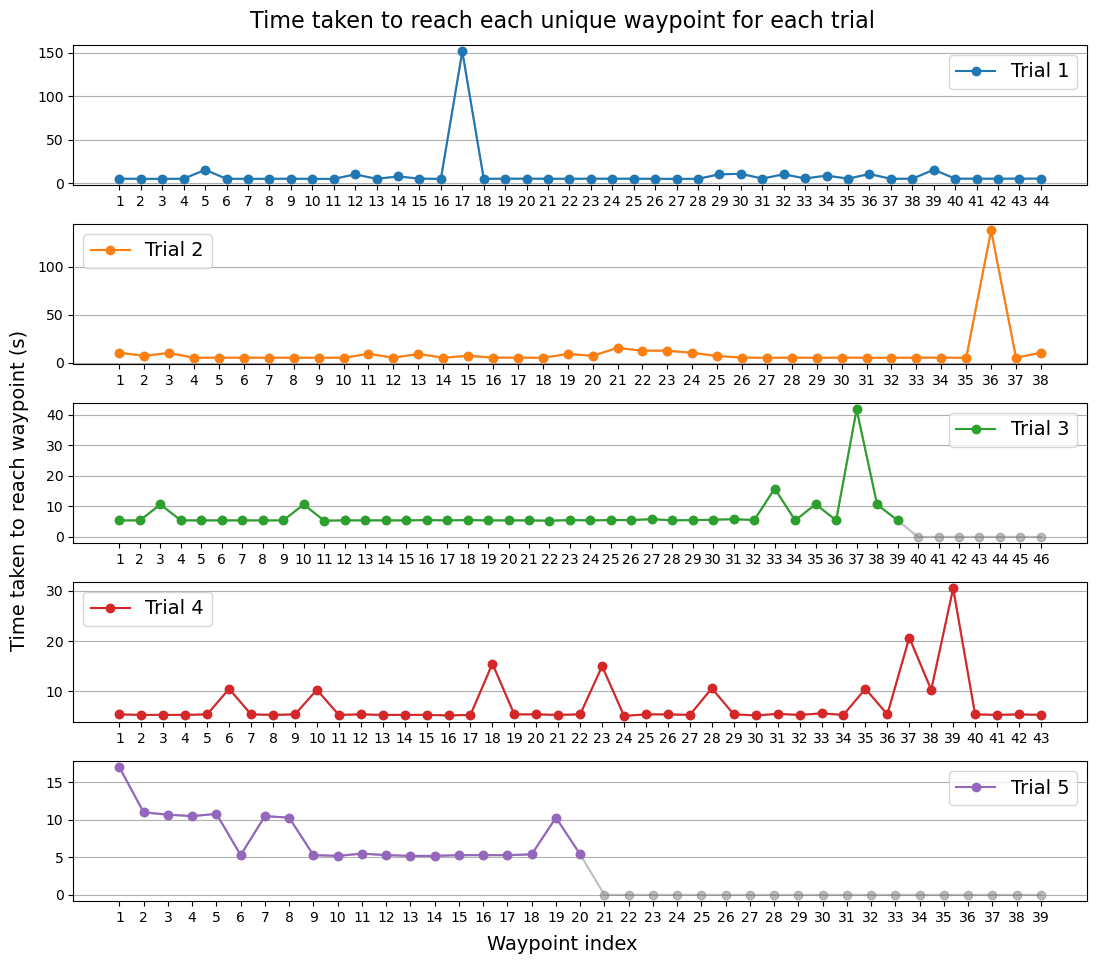}
    \caption{The time taken in seconds for the robot to reach each possible unique waypoint over five trials. Gray dots describe waypoints that the robot has not been able to reach. The outliers occurred due to Nav2 getting stuck at replanning, such that the navigation terminates within narrow desired tolerances, or due to map drift.}
    \label{fig:navigation_time_result}
\end{figure*}

To determine the robot's navigational performance, we evaluate it according to the two aforementioned metrics. This can be seen in Table~\ref{tab:navigation_performance}. In addition, in Figure~
\ref{fig:navigation_time_result}, the time taken for each waypoint over all trials can be seen. We can see that the robot is able to reach $86.5$\% of the total waypoints over all trials with a median time taken of $5.38$ s per waypoint. As seen in Figure~\ref{fig:navigation_time_result}, this was done in the average time of $8.525$ s with a standard deviation of $11.4$ s. 

We can see several outliers where the robot took ranging from 10.0 seconds to a few minutes to complete the navigation to the adjacent waypoint, for instance, at Waypoint 17 in Trial 1. This can be explained due to Nav2 trying to create a local path that neatly ends within the desired set tolerances in position and orientation, and the fact that the robot requires a minimal input velocity in order to move, it overshoots beyond the tolerance and Nav2 has to replan the local path. This can result in the robot being stuck indefinitely, and requires assistance from a human operator. This means that for some of these outliers, ranging from 30 s, human assistance is eventually brought into action. Such that the robot is correctly placed in the desired position and orientation within the set tolerance.

\begin{figure}[tb]
    \centering
    \includegraphics[width=0.75\linewidth]{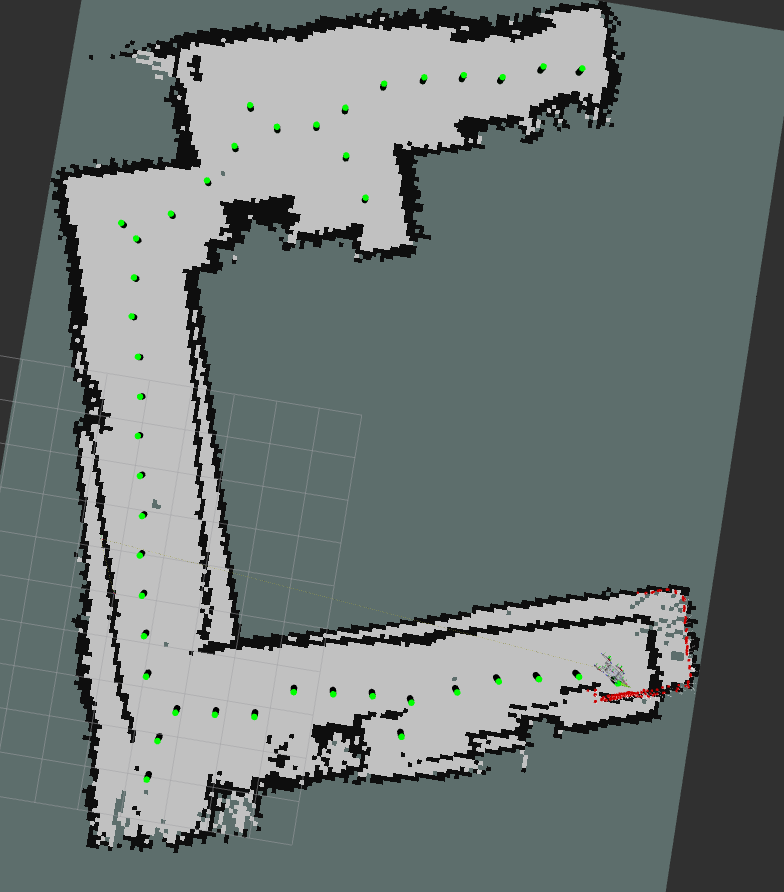}
    \caption{A drift in the map where the hall is doubled. The drift occurred after 10 minutes after the autonomous navigation started in Trial 3.}
    \label{fig:map_drift}
\end{figure}

Furthermore, we can see in Table~\ref{tab:navigation_performance} that in Trial 3 and 5, the robot failed to reach all of the waypoints. This is mainly due to map drifts that occurs over time and the fact that the SLAM module was still operating by mapping the robot's surroundings online. This can be recognized e.g. as two identical hallways slightly tilted from one and another, as seen in Figure~\ref{fig:map_drift}. Due to the drift in the navigation map, since the coordinates of the waypoints remain the same, SLAM adjusts the 2D navigation map such that the later waypoints appear in places that the robot cannot reach, in the walls for example. Drift usually occurs due to the fact that we only use relative sensors to localize itself with respect to the surroundings, e.g. IMU, leg encoders, LIDAR. This can cause the uncertainty to grow over time which is inherent in odometry.

\section{Conclusion}
\label{chapter:conclusion}

Coverage path planning allows mobile robots such as quadrupeds to explore the whole environment. This is especially useful for applications such as surveillance, inspection, and search and rescue. Nevertheless, limited work has been done on autonomous navigation using coverage path planning on quadrupeds. Therefore, we developed an open-source framework using ROS 2 that enables a Unitree Robotics Go2 quadruped to navigate such that it visits every corner using the prior 2D navigation map, autonomously. It utilizes a map reader to extract a graph of 2D waypoints using the topological skeleton of the map, and a path planner to create an efficient path with respect to time and the starting position. A state machine is used to iterate over the ordered list of waypoints and navigate them in succession.

The map reader and the path planner can quickly process maps of width and height ranging between $[196,225]$ pixels and $[185,231]$ pixels in \SI{2.52}{\milli s} and \SI{1.7}{\milli s}, respectively. Their computation time increases with \SI{22.0}{\nano s/pixel} and \SI{8.17}{\micro s/pixel}, respectively. In a closed and unstructured environment, the robot managed to reach 86.5\% of all waypoints over five runs. The failure can be explained due to drifts occurring in the maps over time, because SLAM still operates online. Map drifts can be mitigated using absolute sensors such as Global Positioning System (GPS), and ultra-wideband anchors. Another issue that our path planning does not take into account is obstacles inside the large free space. This can be mitigated by subtracting the contours of said obstacles from the large free space. Skeletonization will account for the creation of waypoints around these occupied spaces. However, the presence of obstacles may necessitate adjustment to the path planner, as they can give rise to cyclical graphs.

Compared to the state-of-the-art methods that use time-constrained planning \cite{bouman2022planning,ly2023planning}, which do not enable variable time of task execution, our approach is not constrained based on a predefined time. While a time-constrained approach is useful when mission time is predefined, the variable-time approach provides a more adaptable solution when operating with limited knowledge. Unlike alternative state-of-the-art approaches that enable time-variable exploration \cite{niu2024skeleton}, our approach is more computationally efficient due to the use of a morphological technique to extract the topological skeleton of the map. Even though the experimental conditions were different, a rough comparison of the map generation magnitude based on the results from each paper shows an order-of-magnitude difference (milliseconds vs microseconds).

Nevertheless, the proposed method is primarily suited for known and static environments, rendering it unsuitable for real-world applications with irregular and moving obstacles within the environment, and varying elevations. Future work will therefore focus on extending the proposed method to incorporate real-time autonomous coverage exploration in unknown 3D environments, particularly under diverse environmental conditions and in the presence of irregular and dynamic obstacles. Additionally, the influence of variations in key parameters will be systematically investigated. 
Moreover, the study will also investigate and seek to improve drift issues commonly encountered in SLAM. Lastly, the proposed approach will be systematically compared to existing 2D coverage path planning methods to evaluate its relative performance and advantages.

\addcontentsline{toc}{section}{References}
\bibliographystyle{IEEEtran}
\bibliography{references.bib}

\begin{thebibliography}{10}
\providecommand{\url}[1]{#1}
\csname url@samestyle\endcsname
\providecommand{\newblock}{\relax}
\providecommand{\bibinfo}[2]{#2}
\providecommand{\BIBentrySTDinterwordspacing}{\spaceskip=0pt\relax}
\providecommand{\BIBentryALTinterwordstretchfactor}{4}
\providecommand{\BIBentryALTinterwordspacing}{\spaceskip=\fontdimen2\font plus
\BIBentryALTinterwordstretchfactor\fontdimen3\font minus \fontdimen4\font\relax}
\providecommand{\BIBforeignlanguage}[2]{{%
\expandafter\ifx\csname l@#1\endcsname\relax
\typeout{** WARNING: IEEEtran.bst: No hyphenation pattern has been}%
\typeout{** loaded for the language `#1'. Using the pattern for}%
\typeout{** the default language instead.}%
\else
\language=\csname l@#1\endcsname
\fi
#2}}
\providecommand{\BIBdecl}{\relax}
\BIBdecl

\bibitem{chiou2022robot}
M.~Chiou, G.-T. Epsimos, G.~Nikolaou, P.~Pappas, G.~Petousakis, S.~Mühl, and R.~Stolkin, ``Robot-assisted nuclear disaster response: Report and insights from a field exercise,'' in \emph{2022 IEEE/RSJ International Conference on Intelligent Robots and Systems (IROS)}, 2022, pp. 4545--4552.

\bibitem{lin2022integrated}
T.-H. Lin, J.-T. Huang, and A.~Putranto, ``Integrated smart robot with earthquake early warning system for automated inspection and emergency response,'' \emph{Natural hazards}, vol. 110, no.~1, pp. 765--786, 2022.

\bibitem{solmaz2024robust}
S.~Solmaz, P.~Innerwinkler, M.~Wójcik, K.~Tong, E.~Politi, G.~Dimitrakopoulos, P.~Purucker, A.~Höß, B.~W. Schuller, and R.~John, ``{Robust Robotic Search and Rescue in Harsh Environments: An Example and Open Challenges},'' in \emph{{2024 IEEE International Symposium on Robotic and Sensors Environments (ROSE)}}, 2024, pp. 1--8.

\bibitem{parades2021mining}
\BIBentryALTinterwordspacing
D.~Paredes and D.~Fleming-Muñoz, ``Automation and robotics in mining: Jobs, income and inequality implications,'' \emph{The Extractive Industries and Society}, vol.~8, no.~1, pp. 189--193, 2021. [Online]. Available: \url{https://www.sciencedirect.com/science/article/pii/S2214790X21000046}
\BIBentrySTDinterwordspacing

\bibitem{ai2024robotasasensorformingsensingnetwork}
\BIBentryALTinterwordspacing
X.~Ai, C.~Xu, B.~Li, and F.~Xia, ``{Robot-As-A-Sensor: Forming a Sensing Network with Robots for Underground Mining Missions},'' 2024. [Online]. Available: \url{https://arxiv.org/abs/2405.00266}
\BIBentrySTDinterwordspacing

\bibitem{jiang2022space}
\BIBentryALTinterwordspacing
Z.~Jiang, X.~Cao, X.~Huang, H.~Li, and M.~Ceccarelli, ``Progress and development trend of space intelligent robot technology,'' \emph{Space: Science \&amp; Technology}, vol. 2022, 2022. [Online]. Available: \url{https://spj.science.org/doi/abs/10.34133/2022/9832053}
\BIBentrySTDinterwordspacing

\bibitem{candalot2024softgrippingspaceexploration}
\BIBentryALTinterwordspacing
A.~Candalot, M.-M. Hashim, B.~Hickey, M.~Laine, M.~Hunter-Scullion, and K.~Yoshida, ``{Soft Gripping System for Space Exploration Legged Robots},'' 2024. [Online]. Available: \url{https://arxiv.org/abs/2411.05482}
\BIBentrySTDinterwordspacing

\bibitem{arm2019spacebok}
P.~Arm, R.~Zenkl, P.~Barton, L.~Beglinger, A.~Dietsche, L.~Ferrazzini, E.~Hampp, J.~Hinder, C.~Huber, D.~Schaufelberger, F.~Schmitt, B.~Sun, B.~Stolz, H.~Kolvenbach, and M.~Hutter, ``Spacebok: A dynamic legged robot for space exploration,'' in \emph{2019 International Conference on Robotics and Automation (ICRA)}, 2019, pp. 6288--6294.

\bibitem{zebro2025}
R.~T. Rajan, A.~Menicucci, A.~Noroozi, P.~Sachdeva, and C.~Verhoeven, ``Lunar zebro – an autonomous moon rover,'' in \emph{IAF Space Exploration Symposium}, 2024.

\bibitem{miller2020mine}
I.~D. Miller, F.~Cladera, A.~Cowley, S.~S. Shivakumar, E.~S. Lee, L.~Jarin-Lipschitz, A.~Bhat, N.~Rodrigues, A.~Zhou, A.~Cohen, A.~Kulkarni, J.~Laney, C.~J. Taylor, and V.~Kumar, ``Mine tunnel exploration using multiple quadrupedal robots,'' \emph{IEEE Robotics and Automation Letters}, vol.~5, no.~2, pp. 2840--2847, 2020.

\bibitem{chagoya2024data}
J.~Chagoya, S.~Patel, C.~Koduru, A.~Kovarovics, M.~H. Tanveer, and R.~C. Voicu, ``Data collection, heat map generation for crack detection using robotic dog fused with flir sensor,'' in \emph{SoutheastCon 2024}, 2024, pp. 824--829.

\bibitem{chen2022nuclear}
\BIBentryALTinterwordspacing
Z.~Chen, H.~Wu, Y.~Chen, L.~Cheng, and B.~Zhang, ``Patrol robot path planning in nuclear power plant using an interval multi-objective particle swarm optimization algorithm,'' \emph{Applied Soft Computing}, vol. 116, p. 108192, 2022. [Online]. Available: \url{https://www.sciencedirect.com/science/article/pii/S1568494621010371}
\BIBentrySTDinterwordspacing

\bibitem{sharma2024reactors}
\BIBentryALTinterwordspacing
U.~Sharma, U.~S. Medasetti, T.~Deemyad, M.~Mashal, and V.~Yadav, ``Mobile robot for security applications in remotely operated advanced reactors,'' \emph{Applied Sciences}, vol.~14, no.~6, 2024. [Online]. Available: \url{https://www.mdpi.com/2076-3417/14/6/2552}
\BIBentrySTDinterwordspacing

\bibitem{yin2023footholds}
Y.~Yin, Y.~Zhao, Y.~Xiao, and F.~Gao, ``Footholds optimization for legged robots walking on complex terrain,'' \emph{Frontiers of Mechanical Engineering}, vol.~18, no.~2, p.~26, 2023.

\bibitem{nitulescu2016designing}
M.~Nitulescu, M.~Ivanescu, V.~D. Hai~Nguyen, and S.~Manoiu-Olaru, ``Designing the legs of a hexapod robot,'' in \emph{2016 20th International Conference on System Theory, Control and Computing (ICSTCC)}, 2016, pp. 119--124.

\bibitem{fan2024review}
Y.~Fan, Z.~Pei, C.~Wang, M.~Li, Z.~Tang, and Q.~Liu, ``A review of quadruped robots: Structure, control, and autonomous motion,'' \emph{Advanced Intelligent Systems}, p. 2300783, 2024.

\bibitem{chai2022survey}
\BIBentryALTinterwordspacing
H.~Chai, Y.~Li, R.~Song, G.~Zhang, Q.~Zhang, S.~Liu, J.~Hou, Y.~Xin, M.~Yuan, G.~Zhang, and Z.~Yang, ``A survey of the development of quadruped robots: Joint configuration, dynamic locomotion control method and mobile manipulation approach,'' \emph{Biomimetic Intelligence and Robotics}, vol.~2, no.~1, p. 100029, 2022. [Online]. Available: \url{https://www.sciencedirect.com/science/article/pii/S2667379721000292}
\BIBentrySTDinterwordspacing

\bibitem{portela2024anymal}
\BIBentryALTinterwordspacing
T.~Portela, A.~Cramariuc, M.~Mittal, and M.~Hutter, ``Whole-body end-effector pose tracking,'' 2024. [Online]. Available: \url{https://arxiv.org/abs/2409.16048}
\BIBentrySTDinterwordspacing

\bibitem{zimmermann2021spot}
S.~Zimmermann, R.~Poranne, and S.~Coros, ``Go fetch! - dynamic grasps using boston dynamics spot with external robotic arm,'' in \emph{2021 IEEE International Conference on Robotics and Automation (ICRA)}, 2021, pp. 4488--4494.

\bibitem{zhang2021objectdetection}
R.~Zhang, L.~Xu, Z.~Yu, Y.~Shi, C.~Mu, and M.~Xu, ``Deep-irtarget: An automatic target detector in infrared imagery using dual-domain feature extraction and allocation,'' \emph{IEEE Transactions on Multimedia}, vol.~24, pp. 1735--1749, 2022.

\bibitem{zhang2025reidentification}
R.~Zhang, Z.~Cao, Y.~Huang, S.~Yang, L.~Xu, and M.~Xu, ``Visible-infrared person re-identification with real-world label noise,'' \emph{IEEE Transactions on Circuits and Systems for Video Technology}, vol.~35, no.~5, pp. 4857--4869, 2025.

\bibitem{zhang2025benchmark}
R.~Zhang, B.~Yang, L.~Xu, Y.~Huang, X.~Xu, Q.~Zhang, Z.~Jiang, and Y.~Liu, ``A benchmark and frequency compression method for infrared few-shot object detection,'' \emph{IEEE Transactions on Geoscience and Remote Sensing}, vol.~63, pp. 1--11, 2025.

\bibitem{bouman2022planning}
A.~Bouman, J.~Ott, S.-K. Kim, K.~Chen, M.~J. Kochenderfer, B.~Lopez, A.-a. Agha-mohammadi, and J.~Burdick, ``Adaptive coverage path planning for efficient exploration of unknown environments,'' in \emph{2022 IEEE/RSJ International Conference on Intelligent Robots and Systems (IROS)}, 2022, pp. 11\,916--11\,923.

\bibitem{niu2024skeleton}
\BIBentryALTinterwordspacing
H.~Niu, X.~Ji, L.~Zhang, F.~Wen, R.~Ying, and P.~Liu, ``{A Skeleton-Based Topological Planner for Exploration in Complex Unknown Environments},'' 2024. [Online]. Available: \url{https://arxiv.org/abs/2412.13664}
\BIBentrySTDinterwordspacing

\bibitem{wang2024quadrupedgpt}
Y.~Mei, Y.~Wang, S.~Zheng, and Q.~Jin, ``{QuadrupedGPT: Towards a Versatile Quadruped Agent in Open-ended Worlds},'' 2024.

\bibitem{guo2024learning}
J.~Guo, Z.~Wang, and W.~Bai, ``Learning quadrupedal robot locomotion for narrow pipe inspection,'' 2024.

\bibitem{cheng2024}
Y.~Cheng, H.~Liu, G.~Pan, H.~Liu, and L.~Ye, ``Quadruped robot traversing 3d complex environments with limited perception,'' in \emph{2024 IEEE/RSJ International Conference on Intelligent Robots and Systems (IROS)}, 2024, pp. 9074--9081.

\bibitem{ly2023planning}
K.~T. Ly, M.~Munks, W.~Merkt, and I.~Havoutis, ``Asymptotically optimized multi-surface coverage path planning for loco-manipulation in inspection and monitoring,'' in \emph{2023 IEEE 19th International Conference on Automation Science and Engineering (CASE)}, 2023, pp. 1--7.

\bibitem{macenski2022ros2}
\BIBentryALTinterwordspacing
S.~Macenski, T.~Foote, B.~Gerkey, C.~Lalancette, and W.~Woodall, ``Robot operating system 2: Design, architecture, and uses in the wild,'' \emph{Science Robotics}, vol.~7, no.~66, p. eabm6074, 2022. [Online]. Available: \url{https://www.science.org/doi/abs/10.1126/scirobotics.abm6074}
\BIBentrySTDinterwordspacing

\bibitem{stachniss2016slam}
\BIBentryALTinterwordspacing
C.~Stachniss, J.~J. Leonard, and S.~Thrun, \emph{Simultaneous Localization and Mapping}.\hskip 1em plus 0.5em minus 0.4em\relax Cham: Springer International Publishing, 2016, pp. 1153--1176. [Online]. Available: \url{https://doi.org/10.1007/978-3-319-32552-1_46}
\BIBentrySTDinterwordspacing

\bibitem{Macenski2021}
\BIBentryALTinterwordspacing
S.~Macenski and I.~Jambrecic, ``Slam toolbox: Slam for the dynamic world,'' \emph{Journal of Open Source Software}, vol.~6, no.~61, p. 2783, 2021. [Online]. Available: \url{https://doi.org/10.21105/joss.02783}
\BIBentrySTDinterwordspacing

\bibitem{rusu2011pcl}
R.~B. Rusu and S.~Cousins, ``{3D is here: Point Cloud Library (PCL)},'' in \emph{{IEEE International Conference on Robotics and Automation (ICRA)}}, Shanghai, China, May 9-13 2011.

\bibitem{macenski2023survey}
S.~Macenski, T.~Moore, D.~V. Lu, A.~Merzlyakov, and M.~Ferguson, ``{From the desks of ROS maintainers: A survey of modern \& capable mobile robotics algorithms in the robot operating system 2},'' \emph{Robotics and Autonomous Systems}, 2023.

\bibitem{dhaeyer1989gaussian}
\BIBentryALTinterwordspacing
J.~P. D'Haeyer, ``Gaussian filtering of images: A regularization approach,'' \emph{Signal Processing}, vol.~18, no.~2, pp. 169--181, 1989. [Online]. Available: \url{https://www.sciencedirect.com/science/article/pii/0165168489900480}
\BIBentrySTDinterwordspacing

\bibitem{cline1987contour}
\BIBentryALTinterwordspacing
W.~E. Lorensen and H.~E. Cline, \emph{Marching cubes: a high resolution 3D surface construction algorithm}.\hskip 1em plus 0.5em minus 0.4em\relax New York, NY, USA: Association for Computing Machinery, 1998, p. 347–353. [Online]. Available: \url{https://doi.org/10.1145/280811.281026}
\BIBentrySTDinterwordspacing

\bibitem{khosravy2017morphology}
\BIBentryALTinterwordspacing
M.~Khosravy, N.~Gupta, N.~Marina, I.~K. Sethi, and M.~R. Asharif, \emph{Morphological Filters: An Inspiration from Natural Geometrical Erosion and Dilation}.\hskip 1em plus 0.5em minus 0.4em\relax Cham: Springer International Publishing, 2017, pp. 349--379. [Online]. Available: \url{https://doi.org/10.1007/978-3-319-50920-4_14}
\BIBentrySTDinterwordspacing

\bibitem{zhang1984thin}
{T. Y. Zhang and C. Y. Suen}, ``{A fast parallel algorithm for thinning digital patterns},'' vol.~27, no.~3, 1984.

\bibitem{barbehenn1998dijkstra}
M.~Barbehenn, ``A note on the complexity of dijkstra's algorithm for graphs with weighted vertices,'' \emph{IEEE Transactions on Computers}, vol.~47, no.~2, pp. 263--, 1998.

\end{thebibliography}

\appendix

This section describes components implemented in the framework that are out of the scope of this research. In App.~\ref{app:sensing_ambiance}, the hardware and the embedded software to measure the ambiance, and the communication with the Go2 Edu robot are described. App.~\ref{app:extended_ros_interface} elaborates the extended interface to convert the data in Go2 Edu robot from custom Unitree ROS messages to commonly used ROS messages to seamlessly work together with open source ROS tools and libraries. Lastly, App.~\ref{app:scanning_procedure} explains the procedure how the Go2 Edu robot scans its immediate surroundings.

\subsection{Sensing Ambiance}
\label{app:sensing_ambiance}

\begin{figure}[tp]
    \centering
    \includegraphics[width=0.8\linewidth]{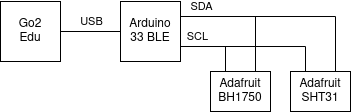}
    \caption{A block diagram illustrating the interconnections between the robot Unitree Robotics Go2 Edu, the microcontroller Arduino 33 BLE, and the sensors Adafruit BH1750 and Adafruit SHT31.}
    \label{fig:sensing_ambiance_diagram}
\end{figure}

Sensing additional parameters of the environment is important for improved awareness, as well as for the robot's usability in various exploration use cases. The robot is equipped with two additional sensors: a sensor \texttt{Adafruit BH1750} that measures the light intensity (in lux), and a sensor that measures both the temperature (in \degree C) and the humidity (in \%) \texttt{Adafruit SHT31}. These sensors are connected to an \texttt{Arduino 33 BLE} micro-controller via a I2C interface. This microcontroller is then connected to the robot via a USB interface. This is illustrated in Fig.~\ref{fig:sensing_ambiance_diagram}.

When the system starts up, it initializes the sensors and begins a serial port with the robot with the baud rate of $115200$ Bd. Afterwards, it periodically\footnote{The measurement rate is arbitrary, here we use $f=5000$ Hz. The minimum frequency should be equal to the sensor that has the highest throttle time.} stimulates the sensors to measure at an instantaneous time. These measurements are quantized as 32-bit floating points. These floats are then transferred to the robot via the serial port with their bit-wise representations. It is important to note that the order is similar between the robot and the microcontroller. This implementation can be seen in \texttt{auxiliary\_sensing} in the root directory.

The ROS 2 node \texttt{auxiliary\_sensors\_node} from the package \texttt{ltm\_go2\_auxiliary\_sensors} creates a CSV-format file and it also connects to the Arduino's serial port upon startup, and translates every incoming packet into their 32-bit floating point representations. This is then published onto the topic \texttt{/ambient\_state}. This information is also added as a line into the CSV-format file with the corresponding timestamp.

\subsection{Extended ROS Interface}
\label{app:extended_ros_interface}

ROS has existing tools and libraries that roboticists can swiftly plug and play into their workspace. One of these libraries is a visualization program called RViz that allows us to see what the robot is doing and perceiving in 3D space.
As mentioned in Section~\ref{chapter:implementation}, the data are only available in Unitree's custom ROS message types whereas RViz uses common ROS message types mentioned in \footnote{Strictly speaking, it is possible to visualize custom message types in RViz if you implement custom RViz plugins. However, this requires additional time and effort.}.

Another library is TF2 which allows us to keep track of multiple reference frames, maintain tree-structured relationships between them using the kinematic model of the robot, and transform points, vectors, etc. from one reference frame to another at any desired moment in time. This is convenient when we want to observe the robot in an inertial reference frame that is not moving along with it. TF2 uses the message type \texttt{geometry\_msgs/TransformStamped} to regularly update the transform between the child reference frame and its parent reference frame.

Besides visualization and transformation, we need to be able to control the robot to perform tasks such as navigation, as well as performing gestures such as sitting down. To fully draw on the available raw data and services, we need an interface to format these data into the corresponding common message types, request services to perform certain tasks, and determine which data we should utilize for visualization and scanning purposes. In addition, to visualize the model of the robot, we use the available URDF of Go2 from \texttt{unitree\_ros} with minor modifications. The following sections describe the interfaces each processing different data types.

\subsubsection{Odometry}
\label{subsection:odometry}

Unitree Go2 has a built-in odometry that defines the robot transformation from some inertial reference frame named \texttt{odom} to the robot's body reference frame named \texttt{base\_link}. Let us denote these as $\mathcal N$ and $\mathcal B$, respectively, and we denote the transformation as ${}^{\mathcal B}C_{\mathcal N}$\footnote{Orientations and rotations are defined in quaternions to avoid the singularity. But, for the sake of clarity, we will be using rotation matrices instead.}. This inertial reference frame is defined as the 3D position where it is initially booted up. 

The odometry data is stored in the \texttt{unitree\_go/SportModeState} message on the topic \texttt{/sportmodestate}. To update the transformation ${}^{\mathcal B}C_{\mathcal N}$, we use position for translation, and quaternion in the message's \texttt{imu\_state} for rotation.

Furthermore, it is also worth publishing the IMU data as a separate entity to record the history. This data is published on the topic \texttt{/imu} using \texttt{sensor\_msgs/Imu} message type. This contains the IMU's measured orientation as \texttt{geometry\_msgs/Quaternion}, angular velocity as \texttt{geometry\_msgs/Vector3}, and linear acceleration as \texttt{geometry\_msgs/Vector3}.

Besides this, another frame is created that is projected from the robot's \texttt{base\_link} frame normal to the ground called \texttt{base\_footprint} $\mathcal{F}$. This is used to project the 2D navigation map onto $z=0$ when $\mathcal N$ as well as the obstacles in the point cloud, as $\mathcal B$ is defined at the robot's height. To obtain for the rotation between \texttt{base\_link} and \texttt{base\_footprint} frames ${}^{\mathcal F}C_{\mathcal B}$, we expand the rotation ${}^{\mathcal F}C_{\mathcal N}$:

\begin{equation}
    {}^{\mathcal F}C_{\mathcal N} = {}^{\mathcal F}C_{\mathcal B}{}^{\mathcal B}C_{\mathcal N} \Rightarrow {}^{\mathcal F}C_{\mathcal B} = {}^{\mathcal F}C_{\mathcal N}{}^{\mathcal B}C^{-1}_{\mathcal N}
\end{equation}

Since \texttt{base\_footprint} is merely a projection of the \texttt{base\_link} onto the ground plane in \texttt{odom} frame, this means that no rotation is applied. Hence ${}^{\mathcal F}C_{\mathcal N}$ is simply an identity matrix $I$. Hence:

\begin{equation}
    {}^{\mathcal F}C_{\mathcal B} ={}^{\mathcal B}C^{-1}_{\mathcal N} 
\end{equation}

The translation $x,y,z$ in $\mathcal F$ depends on the height displacement $z$ in $\mathcal B$, because it is only a vertical projection onto the ground plane. Therefore the translation vector of $\mathcal F$ with respect to $\mathcal B$, $\mathbf x^{F}_{\mathcal B}$, is defined as follows:

\begin{equation}
    \mathbf x_{\mathcal B}^{\mathcal F} =\begin{pmatrix}x_{\mathcal B}^{\mathcal F}\\y_{\mathcal B}^{\mathcal F}\\z_{\mathcal B}^{\mathcal F}\end{pmatrix}={}^{\mathcal F}C_{B}\begin{pmatrix}0\\0\\-z_{\mathcal N}^{\mathcal B}\end{pmatrix}={}^{\mathcal B}C^{-1}_{N}\begin{pmatrix}0\\0\\-z_{\mathcal N}^{\mathcal B}\end{pmatrix}
\end{equation}

\subsubsection{Joint state}
\label{subsection:joint_state}

To visualize the current configuration of the robot on RViz, as well as to update every reference frame on the robot, for instance, when the robot is walking each joint rotates periodically, we would need to extract the angle (joint position) from each of these joints and insert it into a message type specifically designed to track the state of every joint for every measurement. This message type is called \texttt{sensor\_msgs/JointState} which consists of joint names, joint positions, joint velocities, and joint efforts. Note that these are properly ordered, e.g. the first element from each of these lists corresponds to the same joint.

The robot has four legs: front right (FR), front left (FL), rear right (RR), and rear left (RL). Each leg has three joints for every part: hip, thigh, and calf. These joints rotate around the axes $X$, $Y$, $Y$ with respect to the base\_link reference frame (the reference frame placed directly on the pivot center of the robot), respectively.

\subsubsection{Point cloud}
\label{subsection:point_cloud}

The point cloud are previously published as a \texttt{sensor\_msgs/PointCloud2} message on the topic named \texttt{/utlidar/pointcloud} by Unitree. This is measured with respect to the radar frame. 

However, the message's header is not filled which will cause an issue when transforming the point cloud from the radar frame to a desired frame. Therefore, an interface for the point cloud must be created.

This interface acts simply as a passthrough filter while filling the received PointCloud2 message's header with the correct data:

\begin{itemize}
    \item Frame ID: Set as 'radar' to correspond to the correct link name in the URDF.
    \item Timestamp: Unfortunately, we do not have access to the exact time the point cloud was created. So, we set it to the current time it was received.
\end{itemize}

This PointCloud2 message is then published on the topic \texttt{/point\_cloud/raw}.

Additionally, we would also like to request the last received point cloud at any time. This is especially useful for the scanning procedure. Therefore, a service is created named \texttt{get\_pointcloud} which returns the last published point cloud.

\subsubsection{Control}
\label{subsection:control}

The robot is usually controlled manually with a wireless controller. The controller is velocity-based, meaning that depending on the displacement of a joystick the velocity changes accordingly. The output of this controller can be read on the topic \texttt{/wirelesscontroller}.

The robot can also be controlled by publishing messages into this topic with the corresponding message type \texttt{unitree\_go/WirelessController}. It contains four real variables \texttt{lx}, \texttt{ly}, \texttt{rx}, and \texttt{ry}\footnote{It also contains an unsigned integer variable named keys which outputs each button as a binary bit and has corresponding significant position within the variable. '1' is output when pressed, '0' otherwise. However, it is shown that this variable does not do anything using messages.}. \texttt{lx} and \texttt{ly} defines the displacement of the left joystick from its normal position (center), and \texttt{rx} and \texttt{ry} likewise with the right joystick. The following further describes these variables:

\begin{itemize}
    \item \texttt{lx}: Translate the robot laterally, where positive moves the robot to the right, and negative to the left. Normalized in range $(-1,1)$.
    \item \texttt{ly}: Translate the robot longitudinally, where positive moves the robot to the forward, and negative to the backwards. Normalized in range $(-1,1)$.
    \item \texttt{rx}: Rotates the robot about its yaw, where positive rotates the robot clockwise, and negative counter-clockwise. Normalize in range $(-1,1)$.
    \item \texttt{ry}: Rotates the robot about its pitch, where positive tilts the robot to the ground, and negative to the sky. Normalize in range $(-1,1)$.
\end{itemize}

The output of the navigation framework is the desired velocity that the robot must follows in order to traverse the planned trajectory. This output is in \texttt{geometry\_msgs/Twist} which is a message that contains the linear and angular velocities, $\mathbf v_x$ and $\mathbf v_\theta$, respectively. We can map the desired velocity into the controller.

\begin{align}
    l_y &= v_{x,x} \\
    l_x &= -v_{x,y} \\
    r_x &= -v_{\theta,z}
\end{align}

Note that $\mathbf v_x,\mathbf v_\theta$ follows the convention that $x$ always points forward, and the right-hand rule. Furthermore, \texttt{ry} is left untouched as it is not necessary to tilt during navigation, and the fact that the output is only defined on a 2D space.

\subsubsection{Gesture}
\label{subsection:gesture}

Besides the periodic movements such as walking, the robot can also perform customized episodic movements such as standing up, dancing, lying down, etc. We name these types of movements as \textit{gestures}. Besides using the wireless controller, these gestures can only be used behind an API. Therefore, one must request the API in order for the robot the gesture. This can be done by publishing a \texttt{unitree\_api/Request} message with the correct API ID.

Since the LiDAR is pointing towards the ground, we would like the robot to be able to tilt it more towards the sky to achieve a scan of a larger portion of the region of interest. Tilting the robot using \texttt{ry} variable is not sufficient as it only displaces five degrees at most. The robot can perform the gesture to \textit{sit down}, where it tilts the LiDAR farther up. Besides this, we should also be able to make the robot to \textit{stand up} from sitting down, as well as to \textit{lie down} before shutting down or to conserve energy.

Upon further investigation, the robot has two different modes of standing: balanced, and fixed. Balanced standing allows the robot to balance itself by dynamically moving its joints such that its center of mass remains in the margin of stability. Whereas fixed standing ensures that all joints are fixed to specific joint positions such that they appear to be standing still. The former can be achieved using the \textit{Recovery} gesture\footnote{The Recovery gesture also allows the robot to recover from a fall.}, and the latter with the \textit{Standing}.

An interface is implemented to call the gestures with ease using a string as a code. This is done by publishing the code into a \texttt{std\_msgs/String} message on the topic \texttt{/gesture}. Table~\ref{tab:gestures} lists the gestures that are implemented in the interface with their corresponding code in string, and API ID:

\begin{table}[t]
    \centering
    \caption{A list of gestures that are implemented in the extended ROS interface.}
    \label{tab:gestures}
    \begin{tabular}{lcc}
        \toprule
        \textbf{Gesture} & \textbf{Code} & \textbf{API ID} \\
        \midrule
        Fixed standing & \texttt{stand\_up} & $1004$ \\
        Lie down & \texttt{stand\_down} & $1005$ \\
        Balanced standing & \texttt{recovery} & $1006$ \\
        Sit down & \texttt{sit} & $1009$ \\
        \bottomrule
    \end{tabular}
\end{table}

\subsubsection{Camera}
\label{subsection:camera}

The robot is equipped with a 2D camera at the front. It is accessible using OpenCV's video capture method. Note that the camera address and the port must be set correctly, as well as the multi-cast interface address. The latter depends on which hardware is the interface running. If the interface is running locally on the robot, \texttt{eth0} is inserted. If it is running on an external computer via Wi-Fi, then \texttt{wlp2s0} is inserted.

Using \texttt{cv\_bridge}, we can quickly format each captured frame by OpenCV into a \texttt{sensor\_msgs/Image} message. This interface is continuously publishing a sequence of images on the topic \texttt{/camera/raw} at a rate of 24 Hz, however, it is configurable and depends on the hardware's limitation.
In addition to this, the most recent image can be also requested by calling the service \texttt{/get\_image}. This is especially useful for the scanning procedure.

\subsection{Scanning Procedure}
\label{app:scanning_procedure}

The scanning procedure is performed for every waypoint. We are interested in scanning the environment from every angle at a given location as possible. However, this proves to be difficult when the robot can only scan a huge part of the environment when it is sitting. Since sitting down is a discrete motion, this means that it will take a long time when we are scanning for every small angular displacement. It is also worth noting that the robot can only work for several hours as it is running on a battery and depending on the effort. Therefore, we discretize the scanning procedure based on the number of orientations, and the desired gestures.

\begin{algorithm}[t!]
    \begin{algorithmic}[1]
        \State \textbf{Input:} $N \geq 1,\;|G|\geq 1$
        \State Get current robot's 2D pose $x,y,\theta$
        \State $n \gets 0$
        \While{$n \neq N$}
            \For{$g$ in $G$}
                \State Perform gesture $g$
                \State Capture 3D point cloud
                \State Capture 2D image
            \EndFor
            \State $\theta \gets \theta + 2\pi/N$
            \State Navigate towards 2D pose $(x,y,\theta)$
            \State $n \gets n+1$
        \EndWhile
    \end{algorithmic}
    \caption{The pseudocode to perform the scanning procedure at a waypoint.}
    \label{code:scanning_procedure}
\end{algorithm}

Algorithm~\ref{code:scanning_procedure} displays the implemented algorithm to perform the scanning procedure. It takes two inputs: $N$ is the number of orientations, and $G$ is a list of gestures to be performed e.g. standing up, and sitting down. In short, it captures the 3D point cloud and the 2D image in the exact pose for every gesture for each orientation. The 3D point cloud is filtered and transformed into the map frame without the need to assemble fragmented data. This is obtained from the topic \texttt{/point\_cloud/sampled} as illustrated in Figure~\ref{fig:pointcloud_pipeline}.

\end{document}